\documentclass[journal]{IEEEtran}
\usepackage{placeins}
\usepackage{times}
\usepackage{epsfig}
\usepackage{graphicx}
\usepackage{amssymb}
\usepackage{booktabs}
\usepackage{multirow}
\usepackage{verbatim}
\usepackage{cite}
\usepackage{lscape}
\usepackage{tabularx}
\usepackage{amssymb}
\usepackage{hyperref}
\usepackage{gensymb}
\usepackage{subfig}
\usepackage[dvipsnames,svgnames,x11names,table]{xcolor}
\usepackage{caption}
\usepackage{enumitem}
\usepackage{booktabs}
\usepackage{algorithm}
\usepackage{pdfpages}
\usepackage{algpseudocode}
\usepackage{adjustbox}
\usepackage{amsmath} 
\floatname{algorithm}{Algorithm} 
\usepackage{nicefrac, xfrac}

\usepackage{bbm}

\newcommand{\dorowcolors}{\rowcolors{2}{gray!15}{white}}

\newcommand{\mr}[1]{\mathrm{#1}}

\usepackage{xcolor}   
\usepackage{colortbl}

\newcommand{\fImg}{\mathbf{I}}

\newcommand{\xx}{\mathbf{x}}

\newcommand{\data}{\mathcal{D}}

\begin{document}
 \title{\textcolor{black}{Simulations of Common Unsupervised Domain Adaptation Algorithms for Image Classification}}


\author{Ahmad~Chaddad, Yihang~Wu, Yuchen~Jiang, Ahmed~Bouridane, Christian~Desrosiers
\IEEEcompsocitemizethanks{\IEEEcompsocthanksitem A. Chaddad, Y. Wu and Y.Jiang are with the Laboratory for Artificial Intelligence for Personalised Medicine, School of Artificial Intelligence, Guilin University of Electronic Technology, Guilin, China. \\*Corresponding author: Ahmad Chaddad. \\
A. Bouridane is with the Centre for Data Analytics and Cybersecurity (CDAC), University of Sharjah, United Arab Emirates.\\
A. Chaddad and C. Desrosiers are with the Laboratory for Imagery, Vision and Artificial Intelligence, Ecole de Technologie Superieure, Montreal, Canada.\\
Email: ahmad8chaddad@gmail.com, ahmadchaddad@guet.edu.cn}
\thanks{Manuscript received - , 2023; revised  -, 2023.}
}

\maketitle

\begin{abstract}
Traditional machine learning assumes that training and test sets are derived from the same distribution; however, this assumption does not always hold in practical applications. This distribution disparity can lead to severe performance drops when the trained model is used in new data sets. Domain adaptation (DA) is a machine learning technique that aims to address this problem by reducing the differences between domains. This paper presents simulation-based algorithms of recent DA techniques, mainly related to unsupervised domain adaptation (UDA), where labels are available only in the source domain. Our study compares these techniques with public data sets and diverse characteristics, highlighting their respective strengths and drawbacks. For example, Safe Self-Refinement for Transformer-based DA (SSRT) achieved the highest accuracy (91.6\%) in the office-31 data set during our simulations, however, the accuracy dropped to 72.4\% in the Office-Home data set when using limited batch sizes. In addition to improving the reader's comprehension of recent techniques in DA, our study also highlights challenges and upcoming directions for research in this domain. The codes are available at \url{https://github.com/AIPMLab/Domain_Adaptation}.


\end{abstract}

\IEEEpeerreviewmaketitle

\section{Introduction}

\begin{figure}
     \centering
    \includegraphics[width = 0.47 \textwidth]{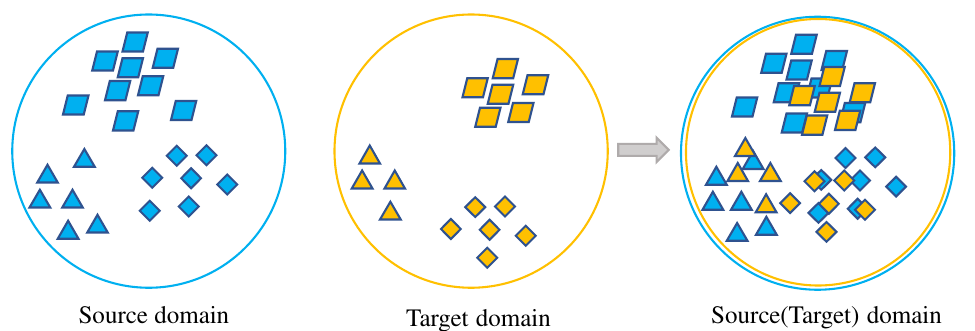}
    \caption{{Illustration of domain adaptation (DA). Left: feature distributions in source and target domains lead to misclassification of target domain samples by a model trained on source data. Right: DA reduces the discrepancy between source and target data distributions, enhancing the model's generalization performance.}}

    \label{F1}
\end{figure}

Rapid advancement in the field of deep learning has led to the emergence of novel approaches that exhibit remarkable performance in various domains, however, achieving high levels of performance with deep learning typically requires a large number of well-annotated training data, which can be challenging to acquire \cite{yang2023auc}, particularly in healthcare. Another problem often encountered in deep learning techniques arises from the fact that the data used to train the model may have different characteristics from the data on which the model is evaluated \cite{chaddad2023domain}. This shift in distribution inevitably leads to a severe drop in performance, as the model cannot generalize well to these new data. In the literature, this problem is commonly referred to as domain adaptation (DA) \cite{chaddad2023enhancing}. Conceptually, DA refers to scenarios where the source and target tasks are identical but the data distributions of the source and target domains differ. In most cases, many labeled samples from the source domain are available, while the target domain has either no labeled samples or only a few. Depending on whether the target domain has labeled data or not, DA can be subdivided into the following three categories: 1) Supervised DA where the target domain data are all labeled, 2) Semi-supervised DA in which part of the target domain data is labeled, and 3) Unsupervised DA where target domain data are without labels. As illustrated in Figure \ref{F1}, applying a model trained on data from the source domain directly to samples from the target domain leads to poor performance. By reducing the discrepancy between domains, DA methods can effectively improve model performance.

DA methods have been proposed for a wide range of applications in computer vision \cite{9972847}. One of the most prevalent applications of DA in this field is image classification \cite{9390356}. For example, the authors of \cite{westfechtel2024gradual} proposed a DA technique that gradually expands the source data with the target data, thus improving the performance model. DA research has also focused on the task of semantic segmentation \cite{zhao2022source}, where the goal is to assign a class label to each pixel in a given image. For example, the work presented in \cite{zhao2022source} introduces a novel source-free open compound DA method for segmentation. This method considers data privacy, multiple target domains, and open, unseen domains. The authors also proposed a cross-patch style swap strategy to enhance the feature-level style of the training samples.

 \begin{figure}
      \centering
    \includegraphics[width = 0.47 \textwidth]{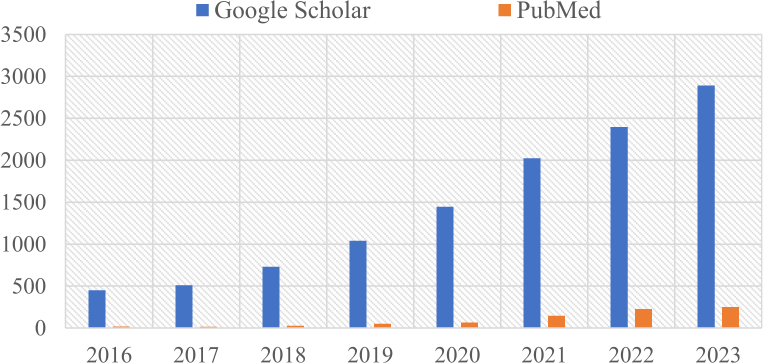}
    \caption{{Bar graph illustrating the number of studies released on the topic of domain adaptation. The data was gathered from searches conducted on Google Scholar and PubMed databases.}}
    \label{F:SearchResult}
\end{figure}

Furthermore, DA has also played an important role in the healthcare field. For example, in \cite{10190316}, they developed unsupervised DA for brain MRI segmentation tasks, achieving higher performance compared to other supervised models. This method greatly reduces the time for manual annotations. In \cite{li2024robust}, they proposed a source-free DA, which uses adversarial sample augmentation and performs self-adaptation to improve the model performance. It shows great value for data privacy protection.

In this paper, we seek to understand the main factors that contribute to the rapid development and success of this field. Figure \ref{F:SearchResult} present the search results for articles related to DA that have been published in leading journals or conferences and the results obtained from the Google Scholar and PubMed databases for the period 2016-2023. {The paper titled "Domain Adaptation" or "Domain adaptive", or "Feature adaptation" was included in the count for both Google Scholar and PubMed.} {This study focuses mainly on unsupervised DA, which is more common and realistic in real-world applications.} {The novelty of our paper lies in an in-depth analysis of commonly used UDA techniques, covering the use of many neural network architectures, datasets, and parameter configurations.} The contributions of our paper can be summarized as follows.
\begin{itemize}
 \item \textcolor{black}{We provide a systematic classification of UDA methods, dividing them into two main categories: traditional methods and deep learning approaches}.
  \item We simulate the most common UDA algorithms for classification tasks using multiple public datasets,  including natural and medical datasets.
  \item We highlight the most commonly used UDA techniques and share our perspectives on selecting optimal neural networks, training approaches, and domain adaptation methods for different types of data. 
    \item We discuss the key challenges and future directions of UDA research for classification tasks.
 \end{itemize}
 
The remainder of this paper is organized as follows. Section \ref{S:2} provides a general definition of DA and discusses previous studies on the topic. Section \ref{S:3} presents the various categories of DA. Section \ref{S:4} describes various commonly used DA methods. Section \ref{S:7} is related to a discussion of various performance metrics for the implementation of DA, including commonly used data sets and the corresponding performance achieved for some applications. Section \ref{S:AB} shows several studies for various DA techniques, along with a detailed analysis. Section \ref{S:9} discusses the current challenges and some potential and future developments and trends of DA. Finally, Section \ref{S:10} gives a set of concluding remarks on the paper.

\section{Background}\label{S:2}

\subsection{The definition of unsupervised domain adaptation}
Given two data domains $\mathcal{D}_s\!=\!\{ {x_i},{y_i}\}_{i = 1}^{{N_s}}$ and $\mathcal{D}_t\!=\!\{ {x_j},{y_j}\}_{j = 1}^{{N_t}}$, where $x$ and $y$ represent samples and labels, respectively. In cases where the feature space and the class space exhibit similarity, but their joint distributions differ (${P_s}(x,y) \ne {P_t}(x,y)$), a common approach is to use the data of the source domain to train a prediction function $f:{x_t} \mapsto {y_t}$ that minimizes prediction loss (denoted by $\ell$) in the target domain \cite{chaddad2023domain}.
\begin{equation}
    {f^*} \, = \, \mathop {\arg \min }\limits_f \ \mathbb{E}_{(x,y) \in \mathcal{D}_t} \big[\ell (f(x),y)\big].
\end{equation}

{However, in UDA, the labels for the target data $y$ are not available during training. The objective is to minimize the prediction loss using labeled source data and unlabeled target data, thereby training a robust classifier that generalizes to the target domain.}

\subsection{Why is domain adaptation developing rapidly}
There exist certain factors that serve as motivators for the implementation of DA. The initial aspect pertains to the inherent characteristics of big data technology, which facilitates the acquisition of vast amounts of data to use in training and updating machine learning processes. However, a significant portion of the data is unique, which requires users to assign appropriate labels to them. To utilize DA effectively, it is advantageous to incorporate labels by identifying data that share a similar labeling with the target data. However, the available training data are constrained, even though numerous applications are filled with complexities and uncertainties. Especially for medical-related applications, accessing large medical data is often considered not practical due to the limited number of cases and privacy leakage problems. DA, in other words, allows one to apply uncontrollable overheads to a broader range of new situations and cases.

\begin{figure*}
      \centering
    \includegraphics[width = 1\linewidth]{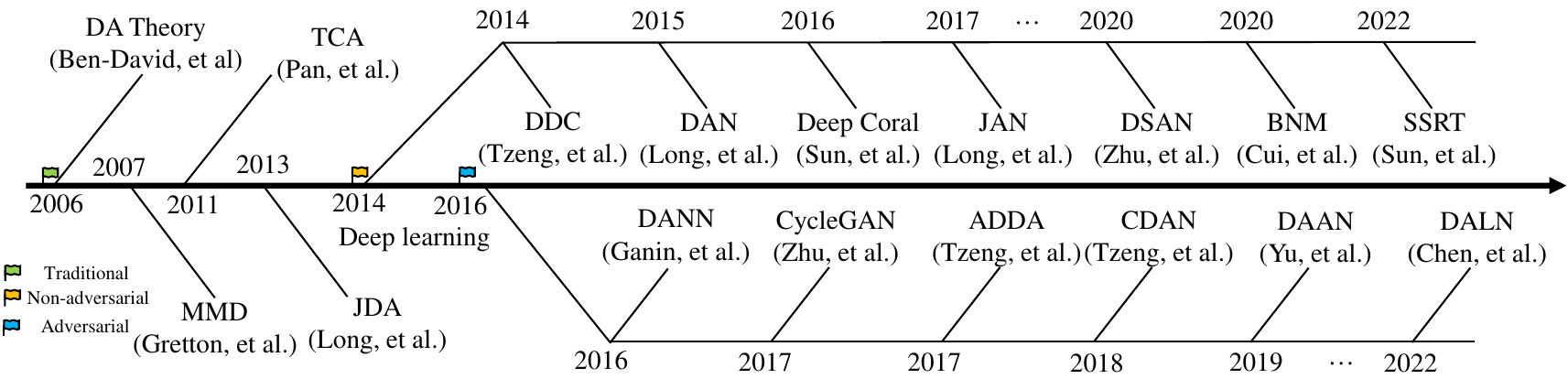}
    \caption{\textcolor{black}{Timeline of domain adaptation techniques. To simplify, this illustration includes common DA methods based on traditional approaches (Green flag) and deep learning techniques (non-adversarial (Orange flag) and adversarial methods (Blue flag)). TCA: Transfer component analysis; MMD: Maximum mean discrepancy; JDA: Joint distribution adaptation; DDC: Deep domain confusion; DANN: Domain adversarial neural network; CycleGAN: Cycle-Consistent Adversarial Networks;DAAN: Dynamic adversarial adaptation networks; DAN: Deep adaptation network; ADDA: Adversarial discriminative domain adaptation; JAN: Joint adaptation network; CDAN: Conditional adversarial domain adaptation; DSAN: Deep subdomain adaptation network; DANN-IB: Generative feature replay for incremental domain adaptation; BNM: Batch nuclear-norm maximization; DALN: Discriminator-free adversarial learning network; SSRT: Safe self refinement for transformer based domain adaptation. }}
    \label{F:Timeline}
\end{figure*}

\section{The Types of Unsupervised Domain Adaptation}
\label{S:3}
\textcolor{black}{This section classifies UDA into two distinct types: traditional UDA, and UDA that uses deep learning methodologies.} \textcolor{black}{In this type of classification, traditional UDA refers to techniques such as TCA, which are more primitive and rely on classical ML methods and hand-crafted/extracted features. Unlike traditional UDA, deep learning-based UDA approaches mainly rely on the dominant feature extraction ability of deep neural networks, further aligning the features of the source and target data directly on fully connected layers, offering better adaptation capabilities and convenience.} We selected papers based on their relevance in the paper title, abstract, and the quality of the paper, such as the number of citations in Google Scholar. \textcolor{black}{Figure \ref{F:Timeline} illustrates the timeline of common traditional and deep learning based DA techniques.} The following Sections, including Section \ref{S:4}, will provide a more detailed explanation of these DA methods.

\subsection{Traditional unsupervised domain adaptation}
The use of traditional DA establishes a robust basis for advancing DA in the future. When considering small datasets, traditional DA is superior to deep neural networks. The present study highlights the prevalent divergence measures used in DA, including MMD \cite{borgwardt2006integrating}, Correlation Alignment (Coral) \cite{sun2016return}, Contrastive Domain Discrepancy (CDD) \cite{kang2019contrastive}. 

{MMD based techniques.} For example, the Transfer Component Analysis (TCA) method, as described in \cite{pan2010domain}, uses the MMD distance metric to evaluate the degree of marginal distribution discrepancy between the source and target domains. MMD, a conventional technique for computing distribution dissimilarity, has undergone several decades of development and has derived some enhanced methodologies. In \cite{long2013transfer}, they introduced a technique called Joint Distribution Adaptation (JDA) that simultaneously adjusts the marginal MMD distance and the conditional MMD distance to improve the adaptation performance. According to \cite{yan2017mind}, existing UDA methods based on MMD do not account for the variations in the prior class distributions, thus contributing to the deterioration of DA performance. The proposal is for a weighted MMD approach that incorporates class-specific weights that account for the prior probability of each class. After related experiments, the results show that the weighted MMD is better than the traditional MMD method. The authors of \cite{zhang2020discriminative} presented a novel discriminative joint probability MMD (JP-MMD) approach to enhance inter-domain transferability by accounting for the discriminability of distinct classes in UDA. In contrast to the conventional MMD approach for computing distribution dissimilarity, JP-MMD enhances transferability and discriminability across diverse domains while being more straightforward and precise.

{Correlation alignment based techniques.} Coral is a conventional technique for domain alignment that primarily attains the linear transformation of the source and target domains to align the second-order features of both domains. On the basis of this method, some new methods are also produced. Deep Coral is one of the very well-known methods. To alleviate the problem of domain mismatch between training and test data sets due to statistical differences, in \cite{li2022coral++}, a new unsupervised DA based on Coral was proposed. In their approach, they improved the Coral algorithm into the Coral++ algorithm for speech recognition. Furthermore, the algorithm has shown excellent results in many tests.

{Contrastive alignment based techniques.} The use of CDD is prevalent in UDA techniques \cite{ding2022source}. The authors of \cite{ding2022source} introduced a novel framework for Source-Free Domain Adaptation (SFDA). The CDD method was incorporated into the framework to enhance intra-class compactness and inter-class separability. Furthermore, the experimental findings indicate that the SFDA approach outperforms conventional UDA techniques using source data.

\subsection{Unsupervised domain adaptation based on deep learning}

\subsubsection{Unsupervised domain adaptation based on adversarial}
Generative Adversarial Network (GAN) was initially introduced as a zero-sum game. Artificial intelligence has recently witnessed a surge in interest towards a particular area, as noted by Goodfellow in his work on generative models \cite{goodfellow2020generative}. The model can be partitioned into two distinct components: the generator and the discriminator. The generator and the discriminator engage in adversarial training to enhance their respective capabilities. The Domain-Adversarial Neural Network (DANN) \cite{ganin2016domain} was the first adversarial UDA. The DANN method takes advantage of the inherent features of the adversarial generative network. The feature extractor and domain discriminator engage in a reciprocal training process to acquire domain-invariant features throughout the training procedure. 

Based on the previous DANN method, more methods are derived from it. \textcolor{black}{For example, inspired by Wasserstein GAN, in \cite{shen2018wasserstein}, they introduce the Wasserstein distance to guide the feature extractor to learn domain-invariant feature representations in an adversarial manner. Empirical studies on natural datasets demonstrate its remarkable performance compared to the state-of-the-art (SOTA) domain-invariant representation learning approaches.} \textcolor{black}{In \cite{pei2018multi}, they demonstrate that using a single domain discriminator is insufficient for fine-grained alignment of different data distributions, further proposing a multi-discriminator adaptation network. Empirical studies indicate that the proposed model outperform SOTA methods with linear-time complexity. Furthermore, in \cite{cao2018partial}, they introduced adversarial DA into partial DA, where the source label space differs from the target label space. Experiments show that it can outperform SOTA results for partial DA.} \textcolor{black}{ In \cite{yu2019transfer}, they extended the DANN to dynamic adversarial adaptation (DAAN). They introduced a balancing parameter to control the impact of the global feature domain classifier and the local feature domain classifier. Experiments demonstrate that their method can improve the performance of DANN in natural image classification tasks.} \textcolor{black}{In \cite{xu2020adversarial}, the authors suggested that domain mix-up \cite{zhang2017mixup} can further guide the domain discriminator in judging the differences of samples relative to the source and target domains. Experimental results using the Office-31 dataset demonstrate that their method can achieve an average accuracy of 81.6\%.}  Furthermore, in \cite{rakshit2022frida}, the authors improved the traditional DANN method and proposed a new DA technique called DANN-IB, which can improve the ability to learn features relevant to the task. Experiments on three benchmark datasets using the DANN-IB framework show that the DANN-IB has better stability than the general method. In \cite{sun2022rethinking}, they rethought adversarial UDA methods, hoping to adequately learn domain-specific representations to alleviate performance degradation caused by subtle domain shifts. So, they propose the orthogonal decomposition adversarial UDA architecture, which can efficiently extract domain-invariant representations. Extensive experiments demonstrate the effectiveness of this method. Furthermore, in a recent work \cite{10025588}, they proposed a novel sparse adversarial domain adaptation model for the classification of traffic scenes. Experimental results in a real-world dataset indicate the impressive results of their method.

\textcolor{black}{Furthermore, in \cite{zhu2017unpaired}, they explored unpaired image-to-image translation tasks with GAN. Unlike paired image-to-image translation, they considered a training set and a test set with different style (e.g., texture shifts), then performed a cycle consistency loss to ensure the training process of GAN. The goal of their approach is to learn a model that can translate an image from a source domain X to a target domain Y in the absence of paired examples. Similarly, in \cite{bousmalis2017unsupervised}, they maintained that models trained purely on synthetic images often fail to generalize to real images, thereby introducing a GAN to perform pixel level adaptation. Experimental results in the USPS dataset indicate that their method can achieve SOTA accuracy.} \textcolor{black}{ In \cite{hoffman2018cycada}, feature space adaptation methods periodically fail to capture pixel-level and low-level domain shifts. Image-space approaches often miss high-level semantic knowledge. They proposed using GAN to perform both pixel-level and feature-level adaptations. The experimental results show competitive accuracy in the classification tasks of digits. \textcolor{black}{In \cite{vu2019advent}, they propose a novel UDA method that performs entropy minimization with adversarial training for semantic segmentation. Experiments using GTA5 and Cityscapes datasets demonstrate the effectiveness of their method in synthetic-2-real setup. In \cite{xia2021adaptive}, they explore the source free UDA with adaptive adversarial network adaptation. The contrastive and self-supervised learning techniques are employed to enhance the target classifier performance. Experiments on Office-31 indicate that their method can achieve an average accuracy of 90.1\% without accessing source data.} In \cite{shi2024adversarial}, they introduced adversarial self-training (AST). AST notably improves both adversarial and clean accuracy in the target domain for gradual DA tasks. They included a case study explaining why AST enhances accuracy.}

However, does GAN necessarily require a special discriminator? In \cite{DALN}, the authors introduced the first discriminator-free adaptive learning network (DALN), which has a good classification effect on some universal datasets in 2022. The method mainly reuses the category classifier as a discriminator again. Recently, the authors of \cite{SUDA} proposed Spectral Unsupervised UDA for Visual Recognition (SUDA) that uses a spectrum transformer to address inter-domain discrepancies. Furthermore, they also proposed a multiview spectral learning procedure that can learn useful feature representations.

\subsubsection{Unsupervised domain adaptation based on non-adversarial}

\textcolor{black}{Similarly to traditional UDA, non-adversarial deep learning based DA can be further grouped into methods based on MMD, correlation alignment, and other approaches.}

{MMD based adaptation. For example, in \cite{tzeng2014deep}, the authors proposed a deep domain confusion (DDC) technique to solve adaptive problems of deep networks for both supervised DA and UDA. Their method mainly used the Alex-net network \cite{krizhevsky2017imagenet} pre-trained on the ImageNet data set for DA learning. Then, they fixed the first seven layers of Alex-net and added adaptive metrics to the $8$th layer, resulting in higher improved performance. In \cite{long2015learning}, they extended the work of DDC, further introduced Deep Adaptation Networks (DAN), a method that simultaneously adds three adaptive layers to constrain the features by adopting the multi-kernel MMD (MK-MMD) metric. Furthermore, condition-based, joint, and dynamic distribution adaptive methods based on MMD have been proposed. For example, the deep subdomain adaptation network (DSAN) \cite{zhu2020deep} is an extensive and flexible distribution adaptation method through a weighted MMD measurement for feature alignment. {DSAN achieves feasible performance on many natural datasets such as Office31 and OfficeHome}. Given the significant performance improvements achieved by the DSAN method over traditional neural network-based DA networks such as DDC and DAN, numerous researchers have shown interest in further enhancing the DSAN model. For example, in \cite{meng2022deep}, they attempted to improve the DSAN method by proposing a new DA method called the Deep Subdomain Associate Adaptation Network (DSAAN) and its application to EEG emotion recognition. The adaptive process of the method is achieved by minimizing the domain classification loss and the Subdomain Associate Loop. Furthermore, several modified methods have shown that DSAAN has remarkable classification ability. In \cite{ran2022cross}, they proposed a modified DSAM method, referred to as the sub-DA feature alignment architecture, to achieve the generalizability of large-scale classification of local climate zones in domains. The sub-DA modules in the framework can effectively reduce the feature distribution differences between domains. \textcolor{black}{Similar to DSAN, in \cite{ge2023unsupervised}, they proposed a conditional MMD based adaptation technique (DCAN) for UDA. They also introduced mutual information loss on target data to train the classifier model.} Furthermore, in \cite{cheng2024deep}, the authors have proposed an adaptation method based on MMD called the Joint Semantic Adaptive Network for multi-source DA. Their proposed Joint Semantic Maximum Mean Discrepancy loss can optimize the cross-domain joint distribution of each class on multiple task-specific layers. Experimental results on the OfficeHome dataset show that their technique can increase accuracy by 2.7\% compared to the SOTA methods.} 

{Deep correlation alignment based methods. Correlation alignment-based methods have seen remarkable advancements. For example, in \cite{sun2016deep}, they proposed that the conventional Coral approach should include feature extraction and linear transformation of the image and training of a support vector machine classifier. This process is known to be laborious. To address this, they integrated Coral with deep learning by developing a loss function that minimizes inter-domain correlation between the source and target domains. In \cite{9774382}, the authors also extend the Coral method by proposing a dynamic joint correlation alignment (DJ-Coral) network for semi-supervised heterogeneous DA. The DJ-Coral network aims to eliminate the heterogeneity of data features between the source and target domains. Experimental results have shown that DJ-Coral offers significant advantages.}

{Other adaptation techniques. A study performed feature alignment using the target features batch nuclear norm in improving learning outcomes in situations lacking labeled data, such as in unsupervised/semi-supervised DA \cite{cui2020towards}. Furthermore, several methods have emerged in recent years based on the same DA approach using BNM. For example, in \cite{cui2021fast}, they proposed an enhancement to existing BNM-based DA methods. Specifically, they have introduced Fast Batch Nuclear-Norm Maximization and Minimization for Robust DA, which aims to improve the prediction accuracy in the target domain. The approach was implemented in three representative scenarios to show the result of the enhancement. Previous studies assume source data can be accessed, however, this may not practical in real world. In \cite{SHOT}, they developed a novel approach called Source Hypothesis Transfer for Unsupervised Domain Adaptation (SHOT), which is one of the pioneering techniques for investigating DA without the need for source data. Furthermore, empirical investigations have shown that SHOT produces cutting-edge results on various DA benchmarks. Furthermore, in \cite{SFDA-2023}, they proposed a self-training approach for SFDA that uses curriculum learning and selective pseudo-labeling to effectively and dependably adapt to domain shifts. This straightforward yet efficient measure prevents the spread of label noise and eliminates the need for costly memory-bank-dependent label refinement. }

The next section will provide a comprehensive analysis of various algorithms, both conventional and innovative, that are frequently employed in the domain of DA. The algorithms that will be discussed include DANN \cite{ganin2016domain}, Deep Coral \cite{sun2016deep}, DSAN \cite{zhu2020deep}, BNM \cite{BNM} and SSRT \cite{SSRT}.

\section{A brief overview of commonly used domain adaptation techniques}\label{S:4}


\subsection{Domain adversarial neural networks}
\begin{figure}[htp]
     \centering
    \includegraphics[width = 0.48 \textwidth]{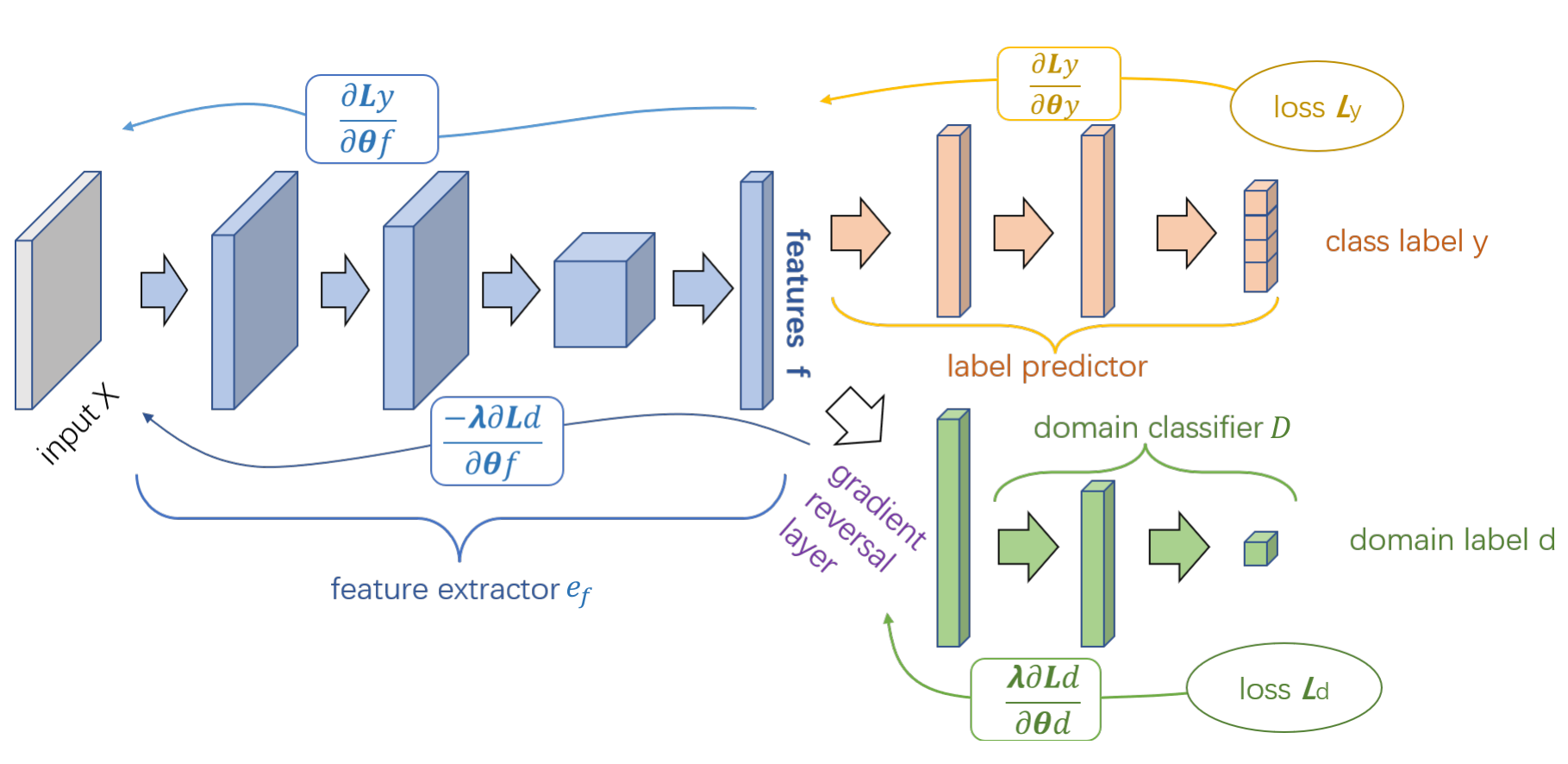}
    \caption{\textcolor{black}{The architecture proposed in \cite{ganin2016domain} comprises a conventional feed-forward design with a label predictor (green) and a feature extractor $e_f$ (blue). The method of unsupervised domain adaptation integrates a domain classifier $D$ (Light salmon), which is connected to the feature extractor $e_f$ via a gradient reversal layer.}}
   
    \label{FDANN}
\end{figure}

Generative Adversarial Networks (GANs) were first introduced into DA in \cite{ganin2016domain}. The technique relies on training a domain classifier to distinguish between data from the source and target domains. In particular, the domain classifier computes the loss and determines whether the data originate from the source or target domain. A gradient inversion layer is added between the domain classifier and the feature extractor during the parameter update phase of backpropagation. The input data are correctly classified into the right domain category (source or target domain) by the domain classifier according to this layer. In addition, they designed a feature extractor to be trained to do the opposite things (due to the gradient inversion layer), to build an adversarial relationship by making the extracted features not accurately classified by the domain discriminator. Figure \ref{FDANN} illustrates the DANN pipeline. {For each iteration, batches of $2B$ samples are randomly generated by selecting the same number of source and target samples. The adversarial DA loss is defined using cross-entropy, as follows:}
\begin{equation}\label{eq:total_loss}
\begin{aligned}
   \mathcal{L}_{DA} \, = \, - \frac{1}{2B} \sum_{j=1}^{2B} z_j\log D_i(\fImg_j)
   \, + \, (1-z_j)\log \big(1-D_i(\fImg_j)\big)
\end{aligned}
\end{equation}

{where $z_j = \mathbbm{1}(\xx_j \in \data_i)$ the domain label of example $\xx_j$. $\textbf{I}_j$ are the features extracted by $e_f$. The domain classifier $D$ predicts if a given image representation $\fImg_j$ is from a \emph{source} ($D_i(\fImg_j) =1$) or a \emph{target} domain ($D_i(\fImg_j) =0$).}

\subsection{Deep Coral}
In \cite{sun2016deep}, they propose that the DA process can be achieved by reducing the discrepancies in higher-order statistical features (e.g., second-order moment features) between the source and target domains. Inspired by the traditional second-order moment feature alignment method (Coral), they propose an end-to-end covariance-based feature alignment method (Deep Coral). It can be incorporated into various strata of Deep Neural Networks and various neural network configurations.

To simplify, suppose that $n_S$ and $n_T$ represent the quantities of source and target data, respectively, with $\mathcal{D}_S^{ij}$ ($\mathcal{D}_T^{ij}$) denoting the value in the j-th dimension of the i-th data sample from the source (target) domain. Assume $C_S$ and $C_T$ denote the covariance matrices for the source and target domain, their Deep Coral metric can be expressed as follows.

\begin{equation}\footnotesize
\ell_{\mr{Coral}}=\frac{1}{4 d^2}\left\|C_S-C_T\right\|_F^2
\end{equation}

where $|\cdot|_F^2$ is the square of the Frobenius norm of a matrix. The dimensionality of the matrix is indicated by the variable $d$. Specifically, the covariance matrices of both source and target data can be measured as follows:
\begin{equation} \footnotesize
\begin{aligned}
&C_S=\frac{1}{n_S-1}\left(\mathcal{D}_S^{T} \mathcal{D}_S-\frac{1}{n_S}\left(\mathbf{1}^{T} \mathcal{D}_S\right)^{T}\left(\mathbf{1}^{T} \mathcal{D}_S\right)\right)
\end{aligned}
\end{equation}
\begin{equation}\footnotesize
\begin{aligned}
&C_T=\frac{1}{n_T-1}\left(\mathcal{D}_T^{T} \mathcal{D}_T-\frac{1}{n_T}\left(\mathbf{1}^{T} \mathcal{D}_T\right)^{T}\left(\mathbf{1}^{T} \mathcal{D}_T\right)\right)
\end{aligned}
\end{equation}

The final loss of deep Coral is expressed as follows.

\begin{equation}
    \ell_{\mr{loss}} \, = \, \ell_{\mr{cla}}+\ell_{\mr{Coral}}
\end{equation}
where $\ell_{\mr{cla}}$ is the classification loss.

\subsection{Deep subdomain adaptation network}
Categorizing similar samples into distinct subdomains within a larger domain, guided by various criteria, such as class labels, has been proposed in \cite{zhu2020deep}. This strategy results in subdomains that cover identical classes, and this particular study uses class categories as the basis for segmentation. Instead of focusing on a comprehensive alignment throughout the domain, this approach emphasizes the independent matching of these localized subdomains.
 
The core of DSAN is to use the local maximum mean discrepancy (LMMD) for DA instead of MMD. Typically, MMD focuses only on feature alignment in the whole dataset, but for many datasets with imbalanced classes, the classes with fewer samples may be incorrectly aligned within the categories with more samples, leading to incorrect DA effects. LMMD, which considers the impacts of each class by introducing weights to different classes, can effectively eliminate this problem. Suppose $\mathcal{H}$ is the reproducing kernel Hilbert space (RKHS) endowed with a characteristic kernel \cite{chaddad2023domain}. $\phi(\cdot)$ is a mapping function that maps the original samples to the RKHS space. The MMD metric can be measured as follows.
\begin{equation}
\hat{d}_{\mathcal{H}}(p, q)=\bigg\|\frac{1}{n_s} \sum_{\mathbf{x}_i \in \mathcal{D}_s} \phi\left(\mathbf{x}_i\right)-\frac{1}{n_t} \sum_{\mathbf{x}_j \in \mathcal{D}_t} \phi\left(\mathbf{x}_j\right)\bigg\|_{\mathcal{H}}^2
\end{equation}

Based on MMD, LMMD has achieved the best results among metric-based methods in recent years, and unlike MMD, it focuses on the discrepancy of local distributions. And their LMMD is defined as:
\begin{equation}
{d_{\cal H}}(p,q) = {{\mathbb{E}}_c}\Big[\left\| {{{\mathbb{E}}_{{p^{(c)}}}}\left[ {\phi \left( {{x^s}} \right)} \right] - {{\mathbb{E}}_{{q^{(c)}}}}\left[ {\phi \left( {{x^t}} \right)} \right]} \right\|_{\cal H}^2\Big]
\label{Eq:29}
\end{equation}
where the variables $x^s$ and $x^t$ refer to the occurrences in datasets $D^s$ and $D^t$, correspondingly. Variables $D^s$ and $D^t$ refer to the source and target domains, respectively. The distributions of $\mathcal{D}_s^{(c)}$ and $\mathcal{D}_t^{(c)}$ are indicated by $p^{(c)}$ and $q^{(c)}$, respectively. An unbiased estimator of Equation \ref{Eq:29} can be expressed as follows:
\begin{equation}\footnotesize
{\hat d_{\cal H}}(p,q) = \frac{1}{C}\sum\limits_{c = 1}^C {\bigg\| {\sum\limits_{x_i^s \in \mathcal{D}_s} {\omega_i^{sc}} \phi \left( {x_i^s} \right) - \sum\limits_{x_j^t \in \mathcal{D}_t} {\omega_j^{tc}} \phi \left( {x_j^t} \right)} \bigg\|_{\cal H}^2}
\label{eq:LMMD}
\end{equation}
where $\omega^c$ is defined according to the number of samples in each category. However, the definition of $\omega$ is very flexible, and this part is easy to extend, such as using multiple similar classes to define sub-fields. 

\subsection{Batch nuclear-norm maximization}
The authors in \cite{BNM} suggest that the ability of a model to distinguish tends to reduce when used with unlabeled data. Thus, they propose the use of entropy constraints as a technique to improve the model's ability to discriminate between classes. However, this approach may result in samples from a few classes being incorrectly classified as the majority class, which may affect the diversity of model predictions. To address this issue, they used F-norm and matrix rank to constrain distinguishability and diversity, respectively. To be more efficient, they propose to use the nuclear norm to approximate the rank of the matrix. They show a lower bound and an upper bound based on the nuclear norm as follows: 

\begin{equation}
    \frac{1}{{\sqrt D }}{\left\| A \right\|_*} \le {\left\| A \right\|_F} \le {\left\| A \right\|_*} \le \sqrt D {\left\| A \right\|_F}
    \label{equa:1}
\end{equation}
where $D=min(B,C)$. $B$ and $C$ here are the batch size and the number of classes, respectively. ${\left\| A \right\|_*}$ is the Nuclear-norm of matrix A.



\begin{figure}[htp]
    \centering
    \includegraphics[width =0.485 \textwidth]{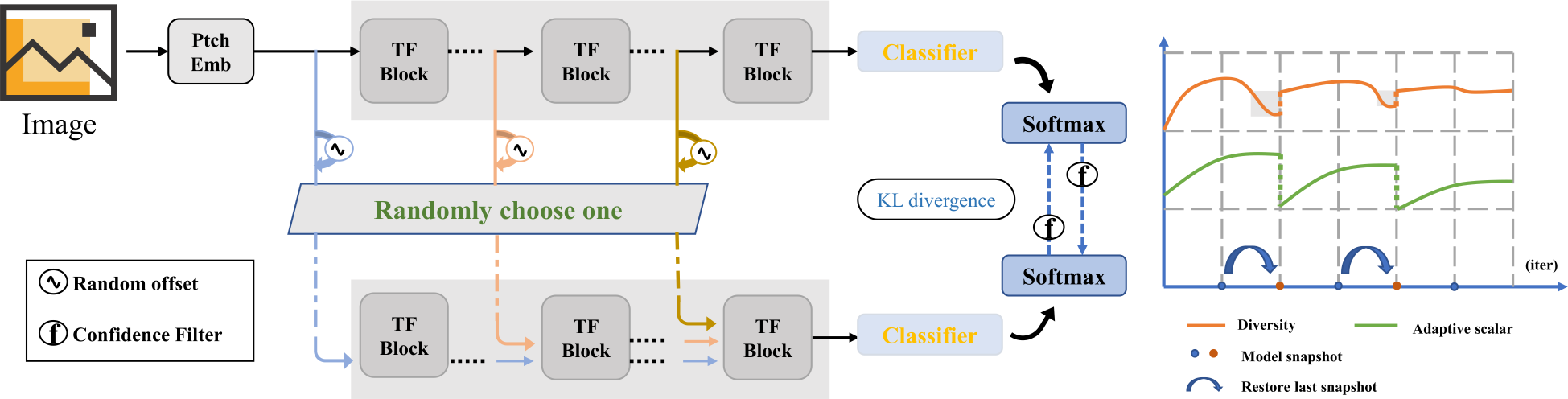}
    \caption{Overview of SSRT \cite{SSRT}. (\textbf{Left}) shows the Self-Refinement for their transformer-based model. Ptch Emb means Patch Embedding, and the two branches share the same parameters. Random offsets are added to the input token sequences of transformer (TF) blocks. The model is trained using its predictions of the original and perturbed versions. It is supervised by Kullback-Leibler divergence (a distance between two probabilities). (\textbf{Right}) shows the Safe Training mechanism.}
 \label{F:SSRT}
\end{figure}

\subsection{Safe self-refinement for transformer-based domain adaptation}
{The SSRT algorithm \cite{SSRT} uses a ViT -- a technique well recognized in the domain of natural language processing \cite{han2022survey} -- to enhance the adaptability of models through judicious predictions on perturbed target domain data \cite{naseer2021generating}}. In particular, perturbations are applied to the hidden label sequence of the target domain data by introducing random offsets. The Kullback-Leibler (KL) divergence is minimized as a measure of the discrepancy between the model's predicted probabilities under the original and perturbed conditions.

The framework of SSRT is illustrated in Fig. \ref{F:SSRT}. It should be noted that only data from the target domain are displayed here. Examining Fig. \ref{F:SSRT} (\textbf{Left}), it can be seen that there are four main components, namely, the TF block, the classifier, the KL divergence, and the Ptch Emb. The TF block is a ViT that is utilized as the entire network. The classifier is utilized for image categorization. Regarding Ptch Emb, the patch Embedding layer transforms the target domain data into a sequence of tokens, which comprise a specific category token and image tokens. Subsequently, they employed TF Blocks to refine this progression. The classifier processes the category token and produces a label prediction. About its random offset, assume that $x$ represents a target domain image and $b_x^l$ is a latent representation of $x$ in a hidden space. Instead of using only $b_x^l$ to perturb the token sequence, they tend to use another target domain image $x_r$ to better perturb the token sequence, which can be represented as:
\begin{equation}
    \bar b_x^l = b_x^l + \lambda {[b_{xr}^l - b_x^l]_ \times }
    \label{SSRT:2}
\end{equation}
where $\lambda$ is a scalar and $[\cdot]_{\times}$ means no back-propagation.

The Kullback-Leibler divergence \cite{joyce2011kullback} measures the dissimilarity between the distributions of samples from the source and target domains. The KL divergence is defined as:
\begin{equation}
    {D_{KL}}\big({p_t}||{p_s}\big) = \sum\limits_i {{p_t}(i)\log \frac{{{p_t}(i)}}{{{p_s}(i)}}}
\end{equation}
where $p_t, p_s$ are target sample probability and source sample probability. Notice that the KL distance is asymmetric, which means ${D_{KL}}\left( {{p_x}||{{\tilde p}_x}} \right) \ne {D_{KL}}\left( {{{\tilde p}_x}||{p_x}} \right)$. Based on this, the authors claimed in their paper that using both KL distances concurrently results in a more resilient model:
\begin{equation}
\begin{aligned}
{\ell_{SR}} \, = \, & {\mathbb{E}_{{B_t} \sim \mathcal{D}_t}} \Big[ \omega\, {\mathbb{E}_{x \sim F\left[ {{B_t};p} \right]}}\, {D_{KL}}\left( {{p_x}||{{\tilde p}_x}} \right)\\
 & \quad + (1\!-\!\omega)\,{\mathbb{E}_{x \sim \;F\left[ {{B_t};\tilde p} \right]}}\, {D_{KL}}\left( {{{\tilde p}_x}||{p_x}} \right)\Big] 
\end{aligned}
\label{SSRT:3}
\end{equation}

To obtain reliable results, they adopt a Confidence Filter $F$. To decrease the risk of model collapse, they then proposed a Safe Training mechanism to learn the model. {Their confidence filter $F$ can be defined as:}
\begin{equation}
    F[\mathcal{D};p] = \{x \in \mathcal{D}| max(p_x) > \epsilon \}
\end{equation}
{where $\epsilon$ is a predefined threshold.}

Regarding the Safe Training mechanism, it is expected that when the model begins to crash, the diversity of model predictions will reduce simultaneously. Therefore, the core idea is to find such collapses during the training process. Once it occurs, the learning setting is reset, and the model is restored to a previous success state. To achieve this, the authors proposed that an adaptive scalar $r \in [0,1]$ is used to adjust $\alpha$ and $\beta$, i.e., $\alpha_r=r\alpha$ and $\beta_r=r\beta$ \cite{SSRT}. Furthermore, they propose a fixed period T and divide the training procedure into successive intervals. At the end of each interval, the model snapshot will be saved simultaneously. Then $r$ is defined as:
\begin{equation}
r(t) = \left\{\begin{array}{l}
\sin\big(\frac{\pi }{{2{T_r}}}(t - {t_r})\big),\ t - {t_r} < {T_r}\\
1.0,\ \text{otherwise}
\end{array}\right.
\label{eq:SSRT1} 
\end{equation}
where $t$ represents the current training step. To begin with, they set $Tr = T$ and $tr = 0$. Then $r$ will increase to 1 by taking $T$ steps. At the end of each interval, according to their Safe Training mechanism, if there is no abrupt drop, $r$ remains the same. Otherwise, $tr$ is reset to the current training step $t$, and the model is restored to the last snapshot. Figure \ref{F:SSRT} (\textbf{Right}) shows the Safe Training mechanism. Note that the diversity of droppings is marked with pink areas.

To determine the diversity of dropping, they proposed a diversity of measurement:
\begin{equation}
    div(t;{\mathcal{B}_t}) = unique\_labels(h({\mathcal{B}_t}))
\end{equation}
where $\mathcal{B}_t$ is a batch of each target training samples, $unique\_labels$ is the unique model predicted labels on target domain. 

In the end, the losses of their model can be formulated as:
\begin{equation}
    \ell  \, = \, {\ell_{CE}} \, - \, {\ell_d} \, + \, \beta {\ell_{SR}}
    \label{SSRT:4}
\end{equation}
where the aforementioned equation involves three distinct loss functions, namely the cross-entropy loss on source data denoted by ${\ell_{CE}}$, the adversarial loss denoted by ${\ell_d}$, and the self-refinement loss denoted by ${\ell_{SR}}$. The parameter $\beta$ represents a trade-off between two or more variables.

\begin{table*}[h]\scriptsize
    \setlength{\tabcolsep}{0.27cm}
    \caption{\textcolor{black}{Commonly used domain adaptation datasets (grouped mainly based on applications).}}
    \dorowcolors
    \begin{tabular}{crrrc} 
      \toprule
      \textbf{ } & \textbf{samples} & \textbf{domains} &\textbf{labels}& \textbf{name of each domain} \\
      \midrule
      \multicolumn{5}{c}{\textsc{Handwritten digit recognition}} \\
      \midrule
      MNIST \cite{MNIST}& 70,000 & 2 &10 classes &train,test\\
      USPS \cite{USPS}& 9,298 & 2&10 classes&train,test\\
      SVHN \cite{netzer2011reading}&600,000&2&10 classes&train,test\\
      \midrule
      \multicolumn{5}{c}{\textsc{Image classification}}\\
      \midrule
      COIL-20 \cite{COIL-20}&1,440&2&20 classes&COIL1, COIL2\\
      Office-31 \cite{Office31}& 4,652 & 3 & 31 classes&Amazon,webcam,dslr \\
      Office-Home \cite{OfficeHome}&15,588&4& 65 classes&Art,clipart,product,real\\
      DomainNet \cite{DomainNet}&586,575&6& 345 classes&Clipart,infograph,painting,quickdraw,real,sketch\\
      VisDA-17 \cite{peng2017visda}& 280,157&2&12 classes &Real images and synthetic\\
      CIFAR-series \cite{CIFA10}& 60,000 & - &10/100 classes &Artificial corruptions\\
      Image\_CLEF \cite{CLEF_DA}& 1,800&3&12 classes &Caltech-256, ImageNet 
 ILSVRC2012, Pascal VOC 2012\\
      Adaptiope \cite{Adaptiope}& 36,900 & 3 &123 classes &Synthetic, product, real life\\
      \midrule
\rowcolor{white}      \multicolumn{5}{c}{\textsc{Activity recognition}}\\
      \midrule
      OPPORTUNITY \cite{OPPO}&701,366&5&4 classes&Back, Right Upper Arm, Right Left Arm, Left Upper Arm, Left Lower Arm\\
      DSADS \cite{ASADS}& 2,844,868&5&19 classes &Torso, right arm, left arm, right leg, left leg \\
      PAMAP2 \cite{PAMAP2}& 1,140,000 & 3 &18 classes &Right arm, Torso, ankle\\
      \midrule
\rowcolor{white}      \multicolumn{5}{c}{\textsc{Semantic segmentation}}\\
      \midrule
      SYNTHIA \cite{SYNTHIA}& 2,700 & 15 &13 classes & 4 locations, 5 weather conditions\\
      GTA5-Cityscapes \cite{GTA5,Cityscapes}& 29,966 & 2 &19 classes &Synthetic and real images\\
      ACDC \cite{ACDC} &4,006 &4 &19 classes &Fog, night, rain , snow\\
      \midrule
\rowcolor{white}      \multicolumn{5}{c}{\textsc{Sentiment classification}}\\
      \midrule
      Amazion Reviews \cite{Amazon_Reviews}& $>$340k&5& 2 classes&DVD,Books,electronics,kitchen appliances\\
      \midrule
\rowcolor{white}      \multicolumn{5}{c}{\textsc{Medical image classification}}\\
      \midrule
    NIH-CXR14 \cite{wang2017chestx}&108,948&2&14 classes&train, test \\
    CheXpert \cite{irvin2019chexpert}&224,316&2&14 classes&train, test \\
    MIMIC-CXR \cite{johnson2019mimic}&377,110&2&14 classes&train, test \\
    Chest X-rays \cite{Chest_Xray}& 5,863 & 3 &2 classes&Artificial corruptions\\

\midrule
\rowcolor{white}\multicolumn{5}{c}{\textsc{Medical image Segmentation}}\\
\midrule
      Multi-site Prostate MRI Segmentation \cite{MRI_Prostate_Seg}& 116 & 6 & 3 classes&Clinical centers\\

Brain Tumor Segmentation \cite{menze2014multimodal}&75&4&2 classes&T1, T1CE, T2, FLAIR \\
      Whole Heart Segmentation \cite{zhuang2016multi}&20&2&4 classes&MRI, CT \\
      ADNI \cite{ADNI}&819&4&3 classes &ADNI-1, ADNI-2, ADNI-GO and ADNI-3\\
      \midrule
\rowcolor{white}      \multicolumn{5}{c}{\textsc{Face recognition}}\\
      \midrule
      FERET \cite{FREAT}& 14,051 & 5 & 1,199 classes&fa, fb, fc, dupI, dupII\\
      PIE \cite{PIE}&11,554&13&68 classes&PIE1,PIE2,$\ldots$,PIE13\\
      \bottomrule
      
    \end{tabular}
    \label{F:Datasets}
\end{table*}
\section{Domain Adaptation Models' Performance}
\label{S:7}
This section provides a summary of the metrics typically employed when assessing the performance of DA models. Furthermore, this section reports the performance of image classification models on benchmark datasets. We have provided an in-depth description of the datasets used in Subsection \ref{SS:1}, while the others have been enumerated in Table \ref{F:Datasets}.

\subsection{Metrics for domain adaptation models}
 In image segmentation tasks, Intersection over Union (IoU) and Dice coefficient (Dice) are commonly used metrics \cite{Segmentation_Metrics}. On the other hand, for classification or recognition tasks, Accuracy, Precision, Recall, Specificity, Area under the ROC curve (AUC) and F1 score are commonly used \cite{Cla_Metrics}.

\begin{table*}[h] \scriptsize
    \caption{{Summary of the implementation details of each algorithm.}}
    \dorowcolors
    \setlength{\tabcolsep}{0.55cm}
    \begin{tabular}{ccccccccc}
    \toprule
     Alg. & Backbone & LR & Epoch & BS &M & OS &WD & Iter \\ 
     \midrule
     Deep Coral & ResNet50 & $3 \times 10^{-3}$ & 30 & 16&0.9 &SGD&$5\times 10^{-4}$&200 \\ 
     DANN &ResNet50 &$1 \times 10^{-2}$ &30 &16&0.9&SGD&$1\times 10^{-3}$&200\\
     DAN&ResNet50&$1\times 10^{-2}$&30&16&0.9&SGD&$5\times 10^{-4}$&200 \\
     DSAN&ResNet50 &$1\times 10^{-2}$ &30 &16&0.9&SGD&$5\times 10^{-4}$&200 \\
     BNM &ResNet50 &$1\times 10^{-3}$ &30 &16&0.9&SGD&$5\times 10^{-4}$&200 \\
     DCAN&ResNet50 &$1\times 10^{-2}$ &30 &16&0.9&SGD&$5\times 10^{-4}$&200 \\
     SSRT &ViT &$1\times 10^{-3}$$^\dag$,$4\times 10^{-3}$$^*$ &10$^\dag$,15$^*$ &4&0.9&SGD&$5\times 10^{-4}$&1000 \\
     w/o DA&ResNet50&$1 \times 10^{-2}$&30&16&0.9&SGD&$5 \times 10^{-4}$&200 \\
     \bottomrule
    \end{tabular}\\
    {LR: Learning rate; BS: Batch size; mom: Momentum; ViT: Vision transformer; OS: Optimization strategy; SGD: Stochastic gradient descent; WD: Weight decay; Iter: iteration per epoch. This table is based on the parameters used in our experiments. The SSRT algorithm's learning rate, according to the public code, is denoted by distinct symbols owing to its dependence on the tested dataset ($\dag$ means on Office-31, $*$ means on Office-Home).}
    \label{T:Details_Alg}
\end{table*}

\begin{table*}[h] \scriptsize
    \setlength{\tabcolsep}{0.57mm}
    \caption{Classification accuracy (Difference) on Office-31 and Office-Home for unsupervised domain adaptation. The Difference is measured based on the deviation of our experimental results from the original data.}
    \dorowcolors
    \begin{tabular}{cccccccccccccc} 
      \hline
      \rowcolor{white}   \multicolumn{14}{c}{Office-31} \\
      \hline
      \textbf{Methods} &\multicolumn{2}{c}{ \textbf{A to W} }&\multicolumn{2}{c}{ \textbf{D to W}} &\multicolumn{2}{c}{ \textbf{W to D}} &\multicolumn{2}{c}{ \textbf{A to D}} &\multicolumn{2}{c}{ \textbf{D to A} }&\multicolumn{2}{c}{ 
      \textbf{W to A}} & \textbf{Average}\\
      \hline     
      ResNet50$^{*}$\cite{ResNet}&\multicolumn{2}{c}{ 68.4 }&\multicolumn{2}{c}{96.7}&\multicolumn{2}{c}{ 99.3}&\multicolumn{2}{c}{ 68.9}&\multicolumn{2}{c}{ 62.5}&\multicolumn{2}{c}{ 60.7}& 76.1\\
      DDC$^{*}$\cite{tzeng2014deep} &\multicolumn{2}{c}{ 75.8}&\multicolumn{2}{c}{ 95.0}&\multicolumn{2}{c}{ 98.2} &\multicolumn{2}{c}{77.5} &\multicolumn{2}{c}{67.4} &\multicolumn{2}{c}{64.0} &79.7\\
      Deep Coral$^{*}$\cite{sun2016deep} & \multicolumn{2}{c}{77.7}& \multicolumn{2}{c}{97.6}& \multicolumn{2}{c}{99.7}& \multicolumn{2}{c}{81.1}&\multicolumn{2}{c}{ 64.6}& \multicolumn{2}{c}{64.0 }&80.8\\
      
      DAN$^*$\cite{long2015learning}&\multicolumn{2}{c}{68.5}& \multicolumn{2}{c}{96.0}& \multicolumn{2}{c}{99.0}& \multicolumn{2}{c}{67.0}&\multicolumn{2}{c}{ 54.0}& \multicolumn{2}{c}{53.1}&72.9 \\
      DANN$^{*}$\cite{ganin2016domain} & \multicolumn{2}{c}{82.0}& \multicolumn{2}{c}{96.9}& \multicolumn{2}{c}{99.1}& \multicolumn{2}{c}{79.7}&\multicolumn{2}{c}{ 68.2}& \multicolumn{2}{c}{67.4}& 82.2\\
      DSAN$^{*}$\cite{zhu2020deep}&\multicolumn{2}{c}{93.6} &\multicolumn{2}{c}{98.3}& \multicolumn{2}{c}{{\textbf{100}}}&\multicolumn{2}{c}{ 90.2}& \multicolumn{2}{c}{73.5}&\multicolumn{2}{c}{74.8}& 88.4\\
      BNM$^{*}$\cite{BNM} &\multicolumn{2}{c}{ 91.5}&	\multicolumn{2}{c}{98.5}&	\multicolumn{2}{c}{{\textbf{100}}}&\multicolumn{2}{c}{	90.3}&\multicolumn{2}{c}{	70.9}&\multicolumn{2}{c}{71.6}&	87.1\\
      SSRT-S$^{*}$\cite{SSRT}&\multicolumn{2}{c}{{\underline{95.7}}}&\multicolumn{2}{c}{ {\underline{99.2}} }&\multicolumn{2}{c}{{\textbf{100}}}& \multicolumn{2}{c}{{\underline{95.8}}}&\multicolumn{2}{c}{ 79.2} &\multicolumn{2}{c}{79.9}& {\underline{91.6}} \\
      SSRT-B$^{*}$\cite{SSRT}&\multicolumn{2}{c}{ {\textbf{97.7}}} &\multicolumn{2}{c}{{\underline{99.2}}}& \multicolumn{2}{c}{{\textbf{100}}}& \multicolumn{2}{c}{{\textbf{98.6}}}& \multicolumn{2}{c}{{\underline{83.5}}}& \multicolumn{2}{c}{{\underline{82.2}}}&{\textbf{93.5}} \\
      \midrule
      Deep Coral$^{\dag }$ &\multicolumn{2}{c}{ 78.0\,(+0.3)	}&\multicolumn{2}{c}{97.3\,(-0.3)}&	\multicolumn{2}{c}{99.7\,(0.0)}&	\multicolumn{2}{c}{76.7\,(-4.4)}&	\multicolumn{2}{c}{63.6\,(-1.0)}&	\multicolumn{2}{c}{64.5\,(+0.5)}	&79.9\,(-0.9)\\
      DAN$^\dag$&\multicolumn{2}{c}{83.0\,(+14.5)}&\multicolumn{2}{c}{97.8\,(+1.8)}&\multicolumn{2}{c}{99.8\,(+0.8)}&\multicolumn{2}{c}{84.3\,(+17.3)}&\multicolumn{2}{c}{69.2\,(+15.2)}&\multicolumn{2}{c}{69.7\,(+16.6)}&83.9\,(+11) \\
      DANN$^{\dag }$ &\multicolumn{2}{c}{ 88.5\,(+6.5)}& \multicolumn{2}{c}{96.7\,(-0.2)}& \multicolumn{2}{c}{{\textbf{100}}\,(+0.9)}& \multicolumn{2}{c}{82.7\,(+3.0)}& \multicolumn{2}{c}{69.1\,(+0.9)}& \multicolumn{2}{c}{72.9\,(+5.5)}& 85.0\,(+2.8)\\
      
      DSAN$^{\dag }$ &\multicolumn{2}{c}{ 92.5\,(-1.1)}&	\multicolumn{2}{c}{98.1\,(-0.2)}&	\multicolumn{2}{c}{{\textbf{100}}\,(0.0)}&	\multicolumn{2}{c}{88.1\,(-2.1)}&	\multicolumn{2}{c}{69.9\,(-3.6)}&	\multicolumn{2}{c}{68.9\,(-5.9)}&	86.3\,(-2.1)\\
      
      BNM$^{\dag }$ & \multicolumn{2}{c}{89.8\,(-1.7)}&	\multicolumn{2}{c}{97.8\,(-0.7)}&	\multicolumn{2}{c}{99.7\,(-0.3)}&	\multicolumn{2}{c}{88.5\,(-1.8)}&	\multicolumn{2}{c}{73.1\,(+2.2)}&	\multicolumn{2}{c}{73.7\,(+2.1)}&	87.1\,(0.0)\\
      
      SSRT-S$^{\dag }$&\multicolumn{2}{c}{92.9\,(-2.8)} &\multicolumn{2}{c}{{\textbf{99.2}}\,(0.0)} &\multicolumn{2}{c}{{\underline{99.9}}\,(-0.1)} &\multicolumn{2}{c}{94.5\,(-1.3) }&\multicolumn{2}{c}{82.2\,(+3.0) }&\multicolumn{2}{c}{81.4\,(+1.5)} &{\underline{91.6}}\,(0.0)  \\
        \hline
     \rowcolor{white}   \multicolumn{14}{c}{Office-Home} \\
        \hline
     \rowcolor{white}   & \textbf{A to C} & \textbf{A to P} & \textbf{A to R} & \textbf{C to A} & \textbf{C to P} & 
      \textbf{C to R} & \textbf{P to A} & \textbf{P to C} &\textbf{P to R} &\textbf{R to A} &\textbf{R to C} &\textbf{R to P}&\textbf{Average}\\
      \hline
      ResNet50$^{*}$\cite{ResNet}& 34.9 &50.0 &58.0& 37.4& 41.9& 46.2& 38.5& 31.2& 60.4 &53.9& 41.2& 59.9 &46.1\\
      DANN$^{*}$\cite{ganin2016domain} & 45.6 &59.3& 70.1& 47.0& 58.5& 60.9 &46.1 &43.7& 68.5& 63.2& 51.8& 76.8& 57.6\\
      DSAN$^{*}$\cite{zhu2020deep} & 54.4& 70.8& 75.4& 60.4 &67.8& 68.0& 62.6& 55.9 &78.5 &73.8& 60.6& 83.1& 67.6\\
      SSRT-S$^{*}$\cite{SSRT}&\textbf{75.2}& \textbf{89.0 }&\textbf{91.1}& \textbf{85.1}& \textbf{88.3}& \textbf{89.9 }&\textbf{85.0}& \textbf{74.2}& \textbf{91.3}& \textbf{85.7 }&\textbf{78.6 }&\textbf{91.8}& \textbf{85.4}\\
      BNM$^{*}$\cite{BNM} & 52.3& 73.9& 80.0& 63.3& 72.9 &74.9& 61.7& 49.5 &79.7 &70.5& 53.6 &82.2& 67.9\\
      \midrule
       DANN$^{\dag }$ & 51.5\,(+5.9)&	67.3\,(+8.0)&	74.2\,(+4.1)&	53.2\,(+6.2)&	65.1\,(+6.6)&	65.4\,(+4.5)&	53.2\,(+7.1)&	50.2\,(+6.5)&	75.1\,(+6.6)&	65.4\,(+2.2)&	57.5\,(+5.7)&	79.5\,(+2.7)&	63.1\,(+5.5)\\ 
      
      DSAN$^{\dag }$ & 53.2\,(-1.2)&	68.5\,(-2.3)&	74.8\,(-0.6)&	58.6\,(-2.4)&	66.5\,(-2.3)&	66.1\,(-1.9)&	59.8\,(-2.8)&	52.4\,(-3.5)&	75.8\,(-2.7)&	71.8\,(-2.0)&	58.6\,(-2.0)&	81.1\,(-2.0)&	65.6\,(-2.0)\\
      
      BNM$^{\dag }$ & 50.4\,(-1.9)&	65.8\,(-8.1)&	76.1\,(-3.9)&	58.6\,(-4.7)&	68.8\,(-4.1)&	71.7\,(-3.2)&	59.8\,(-1.9)&	53.5\,(+4.0)&	80.6\,(+0.9)&	72.6\,(+2.1)&	58.9\,(+5.3)&	81.9\,(-0.3)&	66.6\,(-1.3)
      \\
      
      SSRT-S$^{\dag }$&54.7\,(-20.5)& 75.4\,(-13.6)& 80.6\,(-10.5)& 70.5\,(-14.6)& 78.1\,(-10.2)& 81.5\,(-8.4)& 69.6\,(-15.4)& 54.9\,(-19.3)& 83.5\,(-7.8)& 75.6\,(-10.1)& 58.5\,(-20.1)& 85.6\,(-6.2)& 72.4\,(-13.0)  \\
      \hline
      
    \end{tabular}
    {The highest accuracy is denoted by {\textbf{bold}}, whereas {\underline{underlined}} indicates the second-best accuracy. The approaches denoted by $*$ represent the results of their original papers, while the methods marked by $\dag$ represent our experiment reproduced simulations with publicly available code.}
    \label{T:office31}
\end{table*}

\subsection{The benchmark datasets commonly used in domain adaptation} \label{SS:1}
Table \ref{F:Datasets} provides an overview of the data sets frequently used in the domain of DA, which have been classified primarily by their application, including handwritten digit recognition, image classification, activity recognition, semantic segmentation, sentiment classification, speech recognition, medical imaging, and face recognition. However, some of these datasets are outdated (e.g., MNIST, Image\_CLEF), with many approaches such as DSAN and SSRT, achieving more than 98\% accuracy. With the development of neural networks, the outdated datasets are not suitable for evaluating current SOTA DA techniques, while the basic accuracy using deep networks is highly competitive. Thus, we selected widely used DA benchmark datasets and large-scale natural datasets to assess the performance of DA approaches. The following part elaborates on the datasets used for image classification in our basic examination in more detail.

\textbf{Office31} contains 31 image categories and has 4652 images. It consists of three different domains: Amazon (A), Webcam (W) and Dslr (D). For example, the Amazon domain contains on average 90 images per class and 2,817 images in total. And their pictures come from the Amazon website. The DSLR domain contains 498 low-noise high-resolution images (4288×2848). For the Webcam, it has 795 images of low resolution (640×480) \cite{Office31}. We evaluated those methods in the three domains, which resulted in six DA tasks.

\textbf{Office-Home} is a benchmark dataset consisting of 15588 images with 65 classes \cite{OfficeHome}. This dataset has four different domains: Artistic (Ar, n=2427), Clip art (C, n=4365), Product (P, n=4439), and Real-world (R, n=4357) samples. In the same way, all domains were used and 12 DA tasks were constructed with these methods.

\subsection{Quantitative performance of image classification domain adaptation models}
We assess the performance of the algorithms discussed above in the image classification benchmarks. \textcolor{black}{Moreover, we use an MK-MMD based DA method referred to as DAN \cite{long2015learning}, as well as a conditional MMD based DA technique called DCAN \cite{ge2023unsupervised} for comparative analysis.} This study considers the accuracy of classification, the effect of alignment of the representation of features (visualized using the t-SNE \cite{van2008visualizing} technique) as the main metric. Furthermore, this study uses the most commonly used libraries and software, including Python 3.8, PyTorch 1.13.1, and scikit-learn, to implement these DA techniques. The experiments were performed on a Windows 11 operating system, which features an Intel 11900KF CPU with 128 GB of RAM and an RTX 2060 graphics card. 

Table \ref{T:office31} summarizes the accuracy of image classification models in different datasets. For example, ResNet-50 \cite{ResNet}, DAN \cite{long2015learning}, Deep Coral \cite{sun2016deep}, DANN \cite{ganin2016domain}, DSAN \cite{zhu2020deep}, BNM \cite{BNM}, and SSRT \cite{SSRT} were used to perform a comparison. The best model achieved an accuracy of 93.5\% and 85.4\% using the Office-31 and Office-Home dataset, respectively. The outcomes indicated by $*$ are the results presented in the original articles. The values indicated by $\dag$ represent the results of our simulations. The techniques indicated in the table by $\dag$ show the accuracy and the difference between the original articles and our simulations.

For SSRT, due to technical constraints imposed by our computing hardware, we had to reduce the batch size from 32 (in the original code) to 4 in its public version. The SSRT-S approach uses a slight variant of the ViT model, whereas the SSRT-B approach incorporates the basic structure of ViT. The original results of ResNet, Deep Coral, DANN, DSAN, BNM are all derived from \cite{zhu2020deep}. The rest of the results were taken from the original papers. Furthermore, Table \ref{T:Details_Alg} presents a concise overview of the parameters used in our experimental procedures. Based on our empirical analysis, it has been determined that the SSRT experiences a significant deterioration in its performance when processing large data sets with a limited batch size. {This finding suggests that a small batch size may lead to misalignment for target data by reducing the impact of safe training loss $\ell_{SR}$.} {Furthermore, using the effective learning rate adjustment mechanism (this study uses exponential decay $lr = lr \times {\left( {1 + \gamma  \times float(epoch)} \right)^{ - decay}}$, $\gamma $ is the epoch),} the DANN technique has achieved higher performance compared to the original result. The findings of the analysis of the other algorithms, except SSRT and DANN, were within the acceptable range and showed an average accuracy deviation of less than 3\%.

\section{Exploring domain adaptation techniques} \label{S:AB}

{Currently, the most prevalent applications of DA are in image classification, where it has demonstrated feasible performance in many natural and medical datasets \cite{li2024comprehensive,chaddad2023practical}. The results in Table \ref{tab:Different_NN_2} show that DA significantly increases accuracy and effectively minimizes data distribution discrepancies caused by different capturing devices. This study further examined the potential of DA, specifically in image classification tasks, under various conditions.}

\subsection{Experiments using common neural networks}
This study tested these algorithms using common neural networks. First, the neural networks commonly used in DA were ResNet50, ResNet34, and AlexNet. The data set for the experiments is the Office-Home dataset. Table \ref{tab:Different_NN_2} reports the classification results using the Office-Home dataset. All DA algorithms show a substantial performance improvement when using ResNet50 compared to CNN alone. \textcolor{black}{For example, our simulations show that DCAN can improve the ACC with large margins (e.g., $>20\%$ on task R to P) compared with ResNet50 alone, as reported in Table \ref{tab:Different_NN_2}.} Similarly, there is a large increase in performance when the network is upgraded from AlexNet to the ResNet series; however, there is less enhancement when changing from ResNet34 to ResNet50.

\begin{table*}\scriptsize
    \caption{\textcolor{black}{Classification accuracy using neural networks for unsupervised domain adaptation using Office-Home dataset.}}
    \dorowcolors
    \setlength{\tabcolsep}{0.254cm}
    \begin{tabular}{cccccccccccccc}
    \toprule
         Alg.&Backbone & A to C & A to P & A to R & C to A & C to P & C to R & P to A & P to C & P to R & R to A & R to C & R to P \\
         \midrule
         w/o DA&ResNet50 \cite{ResNet} & 34.9 &50.0 &58.0& 37.4& 41.9& 46.2& 38.5& 31.2& 60.4 &53.9& 41.2& 59.9\\
         \midrule
          & ResNet50 &51.6 &68.3&75.2 &55.9&65.2&67.5&55.0&47.2&74.7 &65.8&54.4&79.6\\
        Deep Coral &ResNet34 &50.5 &63.9&72.0&53.2&62.0&64.8&51.9&46.2&71.5&62.8&52.6&76.8 \\
        & AlexNet &31.8&38.1&50.3 &28.5&41.5&42.9 &27.6&30.4&51.0&40.7&37.5&60.2 \\
        \midrule
        &ResNet50&52.5&67.1&75.3&56.1&64.5&67.9&54.9&47.4&74.9&67.9&55.1&79.8  \\
        DAN&ResNet34& 50.3&64.5&72.4&52.9&62.5&64.9&52.2&47.8&72.9&64.7&54.1&78.1   \\
        &AlexNet& 30.4&39.3&50.3&29.2&42.1&44.1&26.9&30.4&51.7&41.3&37.6&60.7 \\
        \midrule
        & ResNet50 &53.2&68.5&74.8&58.6&66.5&66.1&59.8&52.4&75.8&71.8&58.6&81.1 \\
        DSAN &ResNet34 &49.6&65.9&71.6&54.7&61.4&64.1&56.7&53.2&71.4&67.2&58.3&79.3  \\
        & AlexNet &30.3&39.7&50.8&29.7&39.3&43.4 &26.9&30.3&49.2 & 43.6&38.5&62.4  \\
        \midrule
        & ResNet50 &51.5&67.3&74.2&53.2&65.1&65.4&53.2&50.2&75.1&65.4&57.5&79.5 \\
        DANN &ResNet34 &50.8&63.9&72.2&51.9&62.5&63.9&51.0&51.2&72.5 &63.7 &58.9&77.6  \\
        & AlexNet &31.2&38.7&50.7 &26.9&42.2&43.5&26.2&30.8&51.9&40.8&40.5&60.6  \\
        \midrule
        & ResNet50 &50.4&65.8&76.1&58.6&68.8&71.7&59.8&53.5&80.6&72.6&58.9&81.9 \\
        BNM &ResNet34 &47.8&60.8&71.9&59.3&64.0&62.4&60.0&52.4&76.6 &61.6&57.8&79.1  \\
        & AlexNet &29.7&31.2&48.3&32.6 &41.0&44.8 & 28.8&29.3&50.5&45.2 &34.8 &64.3 \\
          \midrule
          &ResNet50&54.4&68.2&74.7&59.8&68.0&67.8&59.8&54.5&76.3&70.9&60.2&82.5  \\
        DCAN&ResNet34&52.1&66.0&72.3&56.7&65.9&66.4&56.8&55.5&74.9&67.9&59.5&80.8 \\
           &AlexNet&30.7&39.0&50.3&30.4&43.5&45.9&28.6&32.3&54.2&43.6&38.5&63.2     \\
        \bottomrule
    \end{tabular}
{Without DA ResNet34 uses pretrained ResNet34 model and mini-batch SGD with a momentum of 0.9, weight decay 5$\times$ $10^{-4}$ and learning rate $10^{-3}$ to generate the results.}    
    \label{tab:Different_NN_2}
\end{table*}

\subsection{Experiments with modern unsupervised DA dataset} \label{SS:Modern}
\begin{figure}[ht!]
    \includegraphics[width = 0.47 \textwidth]{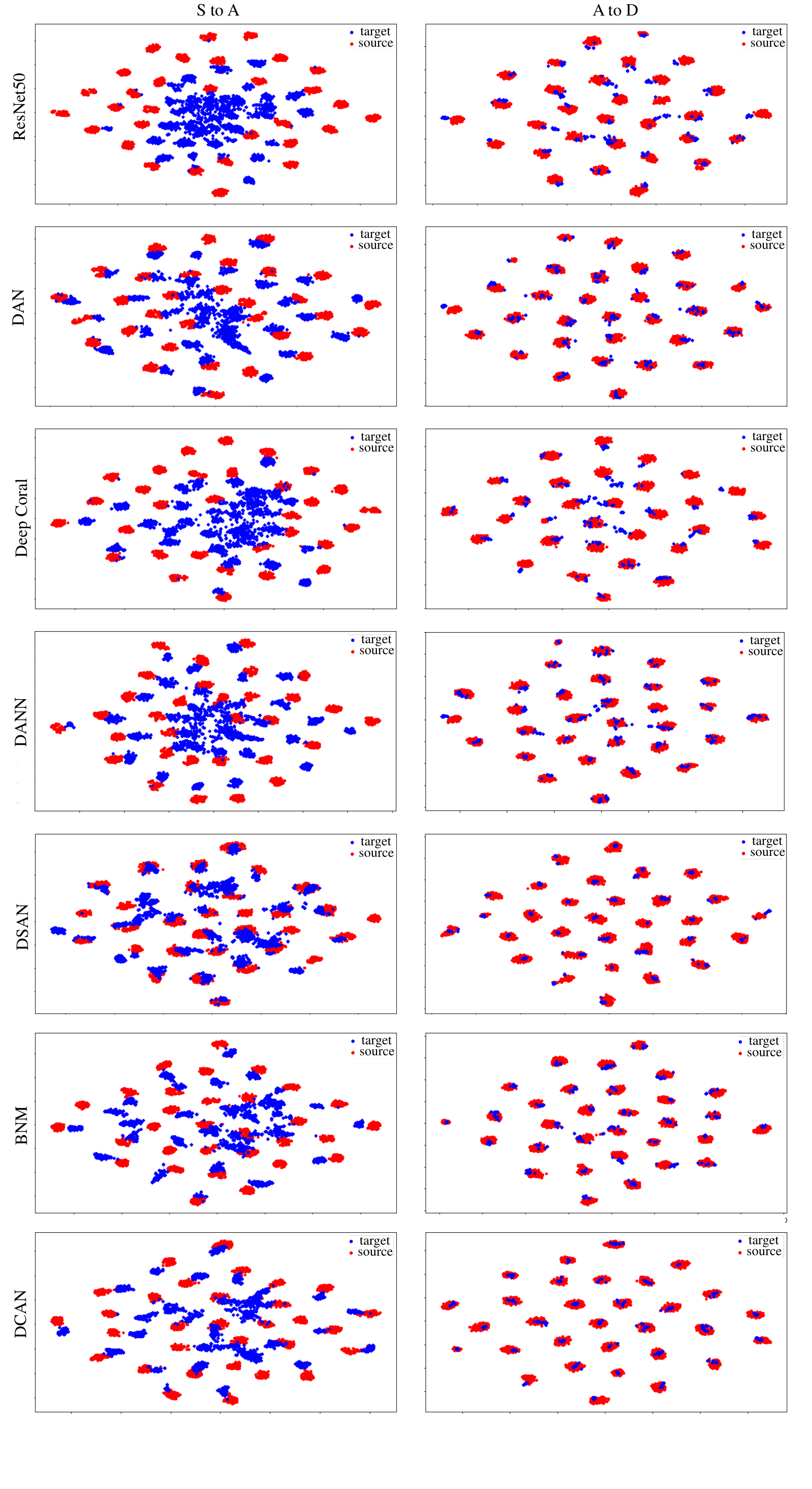}
    \caption{\textcolor{black}{Visualizations of the learned representations using t-SNE for Deep Coral, DANN, DSAN, BNM, DCAN, DAN and ResNet50 alone on task A$\rightarrow$D and S$\rightarrow$A (Modern Office31 dataset), respectively.}}
    \label{fig:ModernDA_TSNE}
\end{figure}

Although these algorithms have remarkable performance using the Office31 data set, the model might use a series of tricks to achieve biased performance. {Thus, this study further uses one more unsupervised DA dataset named Modern Office-31 \cite{ringwald2021adaptiope} to verify these methods.} This dataset has four domains, namely amazon (A), webcam (W), dslr (D) and synthetic (S). {Similarly to Section \ref{SS:1}, four domains were used to build 12 DA tasks}. More details of the implementation can be found in Table \ref{T:Details_Alg}. The DA benchmark test used the following training parameters: mini-batch SGD optimizer with a learning rate of 0.01, batch size of 16, weight decay of 0.0005, momentum of 0.9, 30 rounds of training, and 200 iterations per round. {Table \ref{tab:ModernOffice31} reports the highest test accuracy in the target domain. DANN shows the best performance gain (+9.4\%) compared to ResNet50 alone in ModernOffice31 dataset.} Deep Coral indicates the lowest performance boost (+0.4\%) compared to ResNet50 alone. Surprisingly, DANN outperforms recent DA methods such as DSAN, BNM, which shows the potential to use adversarial methods for DA. Especially for synthesis images, DANN achieves the highest accuracy with a large performance improvement (e.g. $>$ 10\%). Furthermore, Figure \ref{fig:ModernDA_TSNE} shows the representations learned using t-SNE for these techniques on task A$\rightarrow$D and S$\rightarrow$A, respectively. \textcolor{black}{The use of DA leads to better features alignment results compared to ResNet50 alone, particularly using DCAN and DSAN. This highlights the potential of DA for feature adaptation.}

\begin{table*}\scriptsize
    \caption{\textcolor{black}{Classification accuracy (highest) using ResNet50 for unsupervised domain adaptation using modern office31 dataset.}}
    \dorowcolors
    \setlength{\tabcolsep}{0.276cm}
    \begin{tabular}{cccccccccccccc}
    \toprule
         Alg. & A to D & A to S & A to W & D to A & D to S & D to W & S to A & S to D & S to W & W to A & W to D & W to S & Avg \\
         \midrule
        DCAN&92.4&68.4&90.4&84.9&59.6&98.1&82.4&81.7&84.9&84.9&100.0&55.3&81.9\\

        DSAN &92.8 &68.1 &90.4 &84.0 &58.8 &98.9 &81.1 &82.9 &80.9 &84.5 &100 &57.1&81.6  \\
        BNM &88.4 &64.0 &85.2 &84.1 &55.5 &98.4 &84.7 &78.7 &79.1 &87.1 &99.4 &55.7& 80.1  \\
        DANN &87.1 &74.8 &90.9 &87.8 &70.5 &98.5 &88.6 &82.1 &83.3 &88.8 &100 &68.9&85.1  \\
        Deep Coral &85.9 &64.7 &83.4 &78.0 &56.9 &98.7 &72.7 &70.5 &65.4 &79.9 &99.8 &57.7&76.1  \\
        DAN&85.1&65.2&86.0&81.8&61.7&97.8&78.6&82.1&76.2&80.7&99.8&60.2&77.9  \\
        ResNet50 &85.5 &60.1 &82.6 &79.6 &55.7 &98.9 &70.7 &71.1 &67.7 &80.9 &100 &56.1&75.7  \\
        \bottomrule
    \end{tabular}
    \label{tab:ModernOffice31}
\end{table*}

\subsection{Experiments with cross-dataset settings}
Even though these DA techniques demonstrate remarkable performance in standard benchmark datasets, the ability to generalize across various datasets remains a significant challenge. {Therefore, this study selected the commonly occurring categories (backpack, bike, bottle, calculator, desk lamp, file cabinet, keyboard, laptop, monitor, and mouse) from office31 and the office-home data set, to test the generalizability of these DA methods.} Specifically, the study involved the selection of the Amazon (A) and Webcam (W) domains from Office-31, as well as the Art (Ar) and Real World (R) domains from Office-Home for testing. The benchmark test conducted using AlexNet, ResNet34 and ResNet50 without DA used the same training parameters as indicated in Seciton \ref{SS:Modern}. Table \ref{Cross-dataset} shows the classification accuracy using these approaches. {The Deep Coral algorithm exhibited minimal gains in performance using a shallow network such as AlexNet.} Furthermore, when deeper neural networks such as ResNet34 and ResNet50 were used, the algorithm demonstrated limited performance improvement or even negative growth. This leads to the algorithm's limited ability to generalize well when tested on different datasets. Furthermore, the DSAN algorithm shows a performance improvement of 3\%-4\% compared to those without DA. {On the other hand, the DANN algorithm shows a relatively slight improvement using ResNet34, which can be attributed to its limited feature extraction ability to generalize during cross-dataset testing. This suggests that adversarial based techniques may fail to align the feature distributions with inappropriate deep networks.} In terms of the BNM algorithm, when applied to AlexNet, it shows a significant performance improvement of 6.6\%. However, when applied to ResNet, the improvement is comparatively smaller, ranging between 1\% and 4\%. {Furthermore, the use of CNN alone in cross-dataset context shows better performance compared with the results under the same dataset as previously mentioned in Table \ref{tab:Different_NN_2}. The primary reason for this discrepancy is the class scale. OfficeHome and Office31 have 65 and 31 classes, respectively, whereas the cross-dataset has 10. Clustering of data samples on a small class scale is less challenging than under a large scale.}

\begin{table*}[h]\scriptsize
    \caption{\textcolor{black}{Cross-dataset classification accuracy using Office-31 and Office-Home datasets.}}
    \dorowcolors
    \setlength{\tabcolsep}{0.42cm}
    \begin{tabular}{ccccccccccc}
    \toprule
       Alg.&Backbone &A to Ar&A to R&W to Ar&W to R&Ar to A& Ar to W& R to A&R to W&Avg\\
       \midrule
       &AlexNet &52.4&84.8&56.4&78.3&79.7&71.6&85.4&84.1&74.1 \\
        Without DA&ResNet34&67.8&90.8&71.1&89.8&85.2&83.7&91.4&92.8&84.1 \\
          &ResNet50&71.3&94.1&70.6&91.9&86.0&89.8&90.4&94.3&86.1 \\
        \midrule
        &AlexNet&53.6&84.4&57.6&78.7&79.7&68.9&85.0&85.6&74.2 \\
        Deep Coral&ResNet34 &69.4&91.6&73.0&91.9&84.4&84.8&91.5&93.9&85.1 \\
        &ResNet50 &69.0 &93.1&72.5&91.8&84.8&88.3&90.7&94.7&85.6 \\
        \midrule
        &AlexNet&51.8 &83.2 &54.5 &77.7 &76.0 &68.1 &86.7 &86.3 &72.6 \\
        DAN&ResNet34&70.1 &89.4 &67.3 &93.4 &83.3 &79.9 &90.8 &90.5 &81.8 \\
        &ResNet50&69.6 &92.7 &71.3 &92.4 &88.1 &86.7 &91.2&93.9 &85.1  \\
        \midrule
        &AlexNet &59.2 &85.6&60.4&86.5&85.8&73.8&87.6&91.7&78.8 \\
        DSAN&ResNet34 &73.5&96.4&79.1&96.2&85.8&89.0&92.6&94.3&88.4 \\
        &ResNet50 &75.1 &95.0&78.2&96.0&89.3&89.0&93.1&95.1&88.9 \\
        \midrule
        &AlexNet &52.1 &83.8&56.6&84.0&82.2&71.2&86.4&89.0&75.7 \\
        DANN&ResNet34 &67.8 &91.8&73.5&93.0&83.9&82.9&91.1&92.8&84.6 \\
        &ResNet50 &76.5&93.4&72.3&92.6&87.5&88.6&90.8&93.6&86.9 \\
        \midrule
        &AlexNet&61.1&89.5&63.3&88.1&87.8&80.7&88.6&86.7&80.7 \\
        BNM&ResNet34 &66.1 &91.6&68.9&92.9&90.2&88.6&90.0&92.4&85.1 \\
        &ResNet50 &72.3 &96.2&80.3&96.2&91.9&93.6&92.3&95.1&89.7 \\
        \midrule
        &AlexNet&57.1 & 86.9 & 61.6 & 87.5 & 86.2 & 85.2 & 88.0 & 88.6&80.1 \\
        DCAN&ResNet34&76.1 & 95.1 & 77.7 & 95.8 & 88.4 & 92.0 & 92.1 & 95.5&89.1 \\
        &ResNet50&76.8 & 95.3 & 75.4 & 95.3 & 90.2 & 92.4 & 92.7 & 95.5&89.2 \\
       \bottomrule
    \end{tabular}

    \label{Cross-dataset}
\end{table*}

\begin{table*} \scriptsize
    \centering
    \dorowcolors
    \setlength{\tabcolsep}{8pt}
    \caption{\textcolor{black}{Highest classification accuracy using ResNet50 for unsupervised domain adaptation across various data qualities.}}
    \begin{tabular}{cccccccccccc}
    \toprule
        Alg. &$C$ to $C_1$&$C$ to $C_2$&$C$ to $C_3$&$C$ to $C_4$&$C$ to $C_5$& $C_1$ to $C$&$C_2$ to $C$&$C_3$ to $C$&$C_4$ to $C$&$C_5$ to $C$& Avg\\
    \midrule
    Deep Coral&88.2 &84.7 &77.8 &64.5 &41.7&90.0 &87.5 &79.8 &73.7 &58.3 &73.9  \\
    DAN&88.3 &84.5 &77.4 &64.6 &41.4 &91.4 &88.4 &84.5 &73.9 &68.7 &76.3 \\
    DSAN&88.7 &86.1 &79.6 &68.5 &48.9&92.3 &91.0 &87.8 &75.8 &79.2 &81.5 \\
    DANN&89.9 &87.3 &81.3 &70.2 &48.2 &92.7 &90.1 &84.8 &76.3 &76.0 &80.2 \\
    BNM&89.5 &86.6 &80.9 &70.3 &47.7 &90.9 &87.0 &83.7 &75.0 &58.8  &78.3  \\
    DCAN&87.2&84.0&77.3&66.9&44.2&91.7&90.8&80.9&76.5&79.7&77.9\\
    ResNet50&88.1 &84.3 &76.3 &61.3 &38.9 &90.5 &88.3 &79.7 &74.9 &72.0&75.4 \\
    
    \bottomrule
    \end{tabular}

    \label{tab:dataquality}
\end{table*}
\subsection{Experiments under different data quality}\label{S:dataquality}
{The previously mentioned datasets such as OfficeHome and Office31 assume that the data from the source and target domains are of high quality (e.g., without noise or corruption). However, this assumption is impractical in real-world situations. Therefore, we validate the usefulness of DA techniques under different data quality. In particular, ImageNet with different levels of Gaussian noise selected from Imagenet-C \cite{CIFA10} (denoted as $C_1, C_2, C_3, C_4, C_5$) and a tiny ImageNet\footnote{\url{https://www.kaggle.com/datasets/tianbaiyutoby/islvrc-2012-10-pecent-subset}} (denoted as $C$) were used to build 10 DA tasks for evaluation. The total classes used are 100 randomly selected from 1000 classes. The same optimizer and hyperparameter settings are adopted as reported in Table \ref{T:Details_Alg} except the epoch is set to 50. Table \ref{tab:dataquality} reports the highest test accuracy for all DA techniques and CNN alone. When the source data are of high quality, while the target data quality varies, it can be observed that CNN alone indicates a performance drop. However, using DA can increase test accuracy, especially in highly noisy domains such as $C_5$, with an increase in 10\% accuracy using DSAN compared to ResNet50 only. This finding suggests that DA can minimize the shifts in the data distribution when the target data are noisy. Furthermore, when the source data are noisy while the target data are high quality, the potential of DA techniques shows a performance drop when the noise level increase (e.g., $>10\%$ accuracy decrease from $C_4$ to $C$ to $C_5$ to $C$ using Deep Coral). The test accuracy of Deep Coral and BNM in $C_5$ to $C$ is lower compared to ResNet50, which demonstrates negative adaptation.}

\subsection{Comparison with supervised manner}
The studies above are all about UDA. However, the critical role of DA may not be fully explored in a single case. Therefore, following \cite{tzeng2014deep}, 12 DA tasks were constructed using OfficeHome with 10\% samples in the target domain with labels available during training. ResNet50 backbone was adopted for testing. Table \ref{tab:SupervisedDA} reports the test accuracy of these DA techniques. \textcolor{black}{Compared with the UDA results reported in Table \ref{T:office31}, UDA achieves an feasible test accuracy, approximately $\sim 5\%$ lower than the supervised DA. This highlights the ability of UDA.}

\begin{table*}\scriptsize
    \caption{\textcolor{black}{Classification accuracy of ResNet50 for supervised domain adaptation using the Office-Home dataset.}}
    \dorowcolors
    \setlength{\tabcolsep}{8.5pt}
    \begin{tabular}{cccccccccccccc}
    \toprule
         Alg.& A to C & A to P & A to R & C to A & C to P & C to R & P to A & P to C & P to R & R to A & R to C & R to P&Avg \\
         \midrule
          DeepCoral&57.0&72.4&77.6&62.2&69.3&72.0&62.2&54.0&77.6&69.8&58.3&81.4&67.8  \\
          DAN&60.0&74.3&77.5&63.4&71.5&71.2&64.2&57.9&78.1&72.4&63.2&82.7&69.7  \\
        DSAN&59.9&72.8&75.9&63.9&74.0&71.5&66.6&62.0&78.3&74.4&65.5&83.6&70.7  \\
        DANN&58.7&71.8&76.4&61.6&71.0&70.1&62.1&60.3&77.6&71.1&62.7&83.4&68.9 \\
        BNM&56.8&75.4&79.1&66.2&72.8&74.4&65.0&57.0&80.5&72.3&60.9&82.5&70.2 \\
        DCAN&58.7&72.6&76.3&64.6&73.8&70.1&66.1&61.1&78.8&74.8&64.8&83.9&70.5\\
        ResNet50&54.9&69.6&76.1&56.2&67.0&67.2&56.4&48.7&75.1&67.4&56.0&80.7&64.6  \\ 
        \bottomrule
    \end{tabular}
    \label{tab:SupervisedDA}
\end{table*}

\subsection{Experiments with medical datasets}
Medical data can differ between various technologies used in the field, including but not limited to MRI, CT scans, dermoscopic imaging, and Optical Coherence Tomography (OCT). {To test the effectiveness of these DA methods for medical data, the ChestXray17 data set was used, which is related to the classification of pneumonia \cite{kermany2018labeled}.} It consists of 5856 chest radiographs, classified into two groups: pneumonia and normal. Two DA tasks were built for the given dataset, one from the training set to the testing set, and the other from the testing set to the training set. {For both DA and w/o DA, the training epoch is set to 10 rounds and network backbones include AlexNet, ResNet34, and ResNet50, while the rest parameters are the same as reported in Table \ref{T:Details_Alg}.} Table \ref{tab:Medical_1} reports the performance of each algorithm. Using AlexNet, all DA algorithms, except BNM, show a lower performance compared to CNN alone. However, when using ResNet-34 or ResNet-50, Deep Coral and DSAN exceed CNN performance alone.

\begin{table}[h] \scriptsize
    \caption{\textcolor{black}{Experiments using common Domain Adaptation (DA) algorithms using the Chest Xray-17 dataset. Two DA tasks are performed: A and W represent the training and testing domains, respectively}}
    \dorowcolors
    \setlength{\tabcolsep}{0.47cm}
    \begin{tabular}{ccccc}
    \toprule
       Alg.&Backbone &A to B&B to A&Avg\\
       \midrule
       &AlexNet & 93.1&93.2&93.2 \\
        Without DA&ResNet34&93.4&92.6&93.0 \\
          &ResNet50&93.3 &94.0&93.7 \\
        \midrule
        &AlexNet&90.0 &91.1 &90.6 \\
        DAN&ResNet34&87.6 &91.4 &89.5   \\
        &ResNet50&91.3 &92.3 & 91.8  \\
        \midrule
        &AlexNet &91.8&92.2&92.0\\
        Deep Coral&ResNet34 &93.6 &93.7&93.7 \\
        &ResNet50 &94.9 &94.9&94.9 \\
        \midrule
        &AlexNet &92.1 &93.7&92.9\\
        DSAN&ResNet34 &94.4 &93.5&94.0 \\
        &ResNet50 &95.4 &92.5&94.0 \\
        \midrule
        &AlexNet &92.1 &91.8&92.0\\
        DANN&ResNet34 &90.1 &92.2&91.2 \\
        &ResNet50 &91.8 &90.9&91.4\\
        \midrule
        &AlexNet &93.3 &93.9&93.6\\
        BNM&ResNet34 &92.9 &91.4&92.2 \\
        &ResNet50 &94.6 &91.7&93.2 \\
        \midrule
        &AlexNet&90.1&93.7&91.9 \\
        DCAN&ResNet34&91.8&93.6&92.7 \\
        &ResNet50&91.8&92.8&92.3 \\
       \bottomrule
    \end{tabular}
    \label{tab:Medical_1}
\end{table}

\begin{figure}
    \includegraphics[width = 0.47 \textwidth]{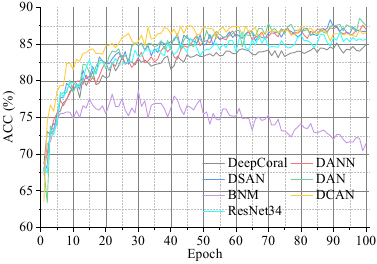}
    \vspace{-3pt}
    \caption{\textcolor{black}{Accuracy of Deep Coral, DANN, DSAN, BNM, DCAN, DAN and ResNet34 alone using the medical dataset.}}
    \label{fig:Medical}
\end{figure}

{Furthermore, this study also performed experiments using multimodal data to evaluate the effectiveness of these DA approaches.} We combine two distinct datasets (Brain tumor, SkinCancer) to evaluate the abilities of these DA strategies. The \textbf{SkinCancer dataset} \cite{tschandl2018ham10000}, comprising 10015 dermatoscopic images classified into seven groups: actinic keratoses and intraepithelial carcinoma / Bowen's disease, basal cell carcinoma, benign keratosis-like lesions, dermatofibroma, melanoma, melanocytic nevi, and vascular lesions. The training dataset consists of 8512 samples, while the test dataset comprises 1503 samples. The \textbf{brain tumor dataset} has four different classes, namely glioma tumor, meningioma tumor, non-tumor and pituitary tumor \cite{BrainTumor_MRI_2}. The training set has 2870 samples, while the testing set has 394 samples. {The model was trained using the training set while test the model on the testing set.} Table \ref{T:Details_Alg} presents a concise overview of the parameters used in our experimental procedures, with 100 epochs for these experiments. Figure \ref{fig:Medical} shows the classification accuracy using these methods. {The use of DA presents a higher average performance and faster convergence compared to using ResNet34 alone. However, BNM results in severe overfitting issues.}

\begin{table*}[t]\scriptsize
    \setlength{\tabcolsep}{2.42mm}
    \renewcommand{\arraystretch}{1.05}
    \caption{\textcolor{black}{The strengths and weaknesses of common and new domain adaptation techniques.}}
    \begin{tabular}{m{7em}  m{5.5cm}  m{5.5cm}  m{2.5cm} m{0.5cm}} 
     \toprule
        \textbf{Method} & \textbf{Strengths} & \textbf{Weaknesses} & \textbf{Datasets}&\textbf{Code}\\
        \midrule
         Deep Coral \cite{sun2016deep}& First integrate Coral alignment with CNN and easy to implement&Poor generalization ability on large-scale datasets & \cite{Office31}&\cite{DDC_Code}  \\
        {DAN\cite{long2015learning}}&{Use multi-kernel MMD to achieve DA, showing better feature adaptation results and higher accuracy}&{It is less suitable for class imbalanced datasets}&{ \cite{Office31,OfficeHome}}&{\cite{DDC_Code}} \\

          DSAN \cite{zhu2020deep} & They choose to align local distribution which can minimize local distribution discrepancies and achieve excellent results & Their defined weight function may not be the best &  \cite{CLEF_DA},  \cite{Office31},  \cite{OfficeHome},  \cite{peng2017visda}&\cite{DDC_Code}\\
          BNM \cite{BNM} &It achieves better discriminability and diversity under label insufficient learning situations& The computational complexity of BNM is approximately $O(n^{3})$, which is time-consuming& \cite{Office31},  \cite{OfficeHome},  \cite{CIFA10}&\cite{BNM}\\
      SSRT \cite{SSRT} &SSRT is among the first to explore ViT for DA&The persistence will decrease in large-scale datasets due to the small batch size scenario& \cite{DomainNet},  \cite{Office31},  \cite{OfficeHome}, \cite{peng2017visda}&\cite{SSRT_Code}\\
        DANN \cite{ganin2016domain} & First introduced GAN to DA problems and achieved state-of-the-art results at that time & No theoretical guarantee for the generalization of DANN to multi-classification problems and regression problems & \cite{MNIST},  \cite{Office31},  \cite{netzer2011reading}&\cite{DDC_Code} \\

        {DCAN \cite{ge2023unsupervised}} &{Remarkable feature adaptation ability, generalizable to multi-modal medical data }& {Time-consuming compared with DANN and DDC} & \cite{Office31}, \cite{OfficeHome}&\cite{DDC_Code}\\
        
         \bottomrule
    \end{tabular}
    {DC: Deep Coral; DA: Domain adaptation; MMD: Maximum mean discrepancy; CNN: Convolutional neural network; ViT: Vision transformer; GAN: Generative adversarial network.}
      \label{T:Summary}
\end{table*}

\subsection{Discussions}
Benefiting from efficient model training methodologies (e.g., early stopping), and advanced hyper-parameter selection techniques, the early DA algorithm DANN was also able to achieve better performance than the original paper's results (e.g., 2.8\% higher average classification accuracy on the Office31 dataset) while using the same neural network. This suggests that the DA algorithm is more effective if an advanced model training approach is adopted. In other words, we claim that if a DA technique shows poor performance, it is preferred to analyze the model training procedure before moving to a different DA method. Additionally, be cautious in choosing the best DA approach while evaluating your experimental setups. The results of our experiments show that the performance of SSRT decreased when a small batch size was used. This observation demonstrates that this approach requires considerable computing power. Moreover, DANN exhibits a high level of promise when employed with a limited batch size.

The most significant improvement in performance can be observed when switching from AlexNet to ResNet series, as shown in Table \ref{tab:Different_NN_2}. However, there is a big difference in the computational resources required by the neural network, ranging from ResNet34 to ResNet50. In low-resource settings, it is critical to select the most suitable neural networks. It is not always helpful to choose deeper networks. For example, if we only run ResNet50 with a batch size of 4, but run ResNet34 with a batch size of 16, the performance will not be the same. {This is because certain DA techniques, such as SSRT \cite{li2020maximum}, require a larger batch size to perform well, as previously mentioned in this study}.

Simulations using a modern office31 dataset have also demonstrated that with the advanced hyperparameter tuning strategy, DA methods can still achieve comparable performance even if they are not recently proposed. This indicates that it is also necessary to consider a suitable training strategy, not only to focus on developing new DA techniques.

In the case of cross-datasets, the performance of deep neural networks like ResNet50 is found to be unsatisfactory compared to using a single dataset like Office31. This suggests that the effectiveness of DA has been reduced due to significant differences in the data sets or due to the influence of data set changes on the performance of DA to some degree. However, this strategy is not practical, as in real-life situations, data can originate from various sources rather than a preset dataset. If a model only shows good performance on a specific dataset, it lacks generality. 

{Furthermore, data quality is important when applying DA techniques. For example, the findings presented in Section \ref{S:dataquality} demonstrate that data quality can affect the adaptation procedure, resulting in a performance drop compared to using CNN alone. In the case of supervised DA, Table \ref{tab:SupervisedDA} indicates that unsupervised DA can achieve closed accuracy with only 3.6\%  drop for BNM under UDA compared to the supervised approach. It highlights the potential of unsupervised DA techniques.} 

Tests performed on medical data sets indicate that some DA approaches that are successful in using natural images may not be applicable to medical data. While BNM displays superior performance on the Office-series dataset, its performance falls short on medical datasets. MMD-based techniques, such as DSAN, demonstrate remarkable efficacy in handling medical data, especially for MRI and dermoscopic images. We suggest that the use of MMD-based techniques could be more effective in addressing distribution discrepancies in medical data. So far, the strengths and weaknesses of the popular and novel methods discussed herein are summarized in Table \ref{T:Summary}.

\section{Challenges and Future Trends that Domain Adaptation Meet}
\label{S:9}

\subsection{Domain adaptation in real world scenarios}
Despite the impressive results obtained from existing clean datasets, such as office series datasets, in the field of DA, it is rare to see such datasets in the real world; with most of the data messy, the application of existing methods to these datasets is likely to produce unsatisfactory results \cite{xu2024low}. {Furthermore, the experimental results with different data quality as reported in Table \ref{tab:dataquality} indicate that the use of DA can improve the performance when the source and target data are noisy.} Similarly, in \cite{noisy_DA}, a scenario was proposed in which the labels were corrupted by noise and then DA was used to address this problem.

\subsection{Domain adaptation under data privacy protection}
Although DA has produced favorable results, most techniques require a substantial amount of data from the source domain. The increasing refinement of legislation and policies related to data privacy regulation, such as the General Data Protection Regulation (GDPR), has resulted in a significant challenge in the use of a large number of data for training purposes \cite{zhang2023federated}. Therefore, it is worthwhile to study how to deal with this challenge. For example, the work of \cite{SHOT} attempts to identify domain-invariant representations that can align the source distribution by maximizing information using self-supervision loss. The work \cite{PDA} attempts to refine the source model by applying entropy functions to generate reliable pseudo-labels for target samples.


\subsection{Domain adaptive based foundation models}
{Recently, vision–language based foundation models (FMs) have demonstrated remarkable performance in many applications, including image classification and segmentation \cite{gao2024clip}. However, the superior feature extraction abilities of FMs in the context of DA have not been fully investigated. In \cite{hu2024reclip}, the authors investigated the potential of the Contrastive Language Image Pre-Training (CLIP) model in source-free domain adaptation (DA). Their experimental results on 23 image classification benchmarks demonstrated the feasibility and efficacy of the proposed approach, which outperformed the SOTA method. Similarly, in \cite{lai2024empowering}, they proposed a novel UDA technique using CLIP for image classification. The experimental results in the OfficeHome dataset indicate that the proposed method outperforms traditional UDA techniques, such as CDTrans \cite{xu2021cdtrans} and TVT-B \cite{yang2023tvt}, with an improvement in accuracy $>$ 7\%. 


\subsection{Self-supervised domain adaptation}
The efficacy of DL can be attributed to the availability of a substantial amount of training data. Occasionally, it may be impractical to collect the requisite amount of accurately annotated data for each undertaking. {The available source training data may not be sufficient. Self-supervised DA can use auxiliary tasks to obtain labels and employ supervised learning techniques to extract informative representations of tasks in the target domain, providing a possible solution \cite{Self_supervised2022}.}

\subsection{Test time adaptation}
{Similarly to source-free DA as illustrated in \cite{SHOT}, test time adaptation (TTA) adapts the model trained in the source domain directly to the target data. This process does not require access to the source data, which is common in the healthcare field. For example, in \cite{chen2022contrastive}, they argued that self-supervised contrastive learning facilitates target feature learning. They also proposed an online pseudo labeling scheme for unsupervised adaptation. Furthermore, in \cite{mounsaveng2024bag}, the authors conducted a comprehensive simulation study using four TTA techniques. This study highlighted the potential of TTA with many settings, such as computational resources and model complexity.}

\subsection{Out of distribution adaptation}
Many existing UDA algorithms rely on a static data distribution, which means that once the dataset is established, the distribution of the data within it remains fixed. However, this assertion frequently fails to hold in practical situations. In practical situations, data tend to be dynamic in nature. For example, satellite maps and face data are subject to constant change over time \cite{ODA}. So, developing some UDA techniques to solve this problem is critical.

\section{Conclusion}\label{S:10}

This study provided a detailed exploration of common UDA techniques. It offered illustrative examples of UDA algorithms and a comprehensive overview of these models. {This study simulates nine experiments to test UDA methods such as Deep Coral, DANN, DSAN, BNM, DCAN, SSRT and DAN.} For example, the SSRT technique delivered remarkable performance when tested on the office31 data set. However, its performance suffered a significant drop when applied to the office-home data set because of the use of a limited batch size. We evaluated both the merits and demerits of the common DA techniques. We also looked at the future trajectories of UDA and addressed the challenges it currently faces.

\section*{Acknowledgments}
This research was funded by the National Natural Science Foundation of China grant number 82260360, the Guilin Innovation Platform and Talent Program 20222C264164, the National Innovation Training Program for College Students under Grant 202310595083, and the Guangxi Science and Technology Base and Talent Project (2022AC18004, 2022AC21040).

{
\bibliographystyle{ieeetr}
\bibliography{reference}
}

\end{document}